\documentclass{article}

\usepackage{microtype}
\usepackage{graphicx}
\usepackage{subcaption}
\usepackage{booktabs} 
\usepackage{hyperref}

\usepackage[preprint]{icml2026}

\usepackage{amsmath}
\usepackage{amssymb}
\usepackage{mathtools}
\usepackage{amsthm}
\usepackage{microtype}      
\usepackage{xcolor}        
\usepackage{amsmath} 
\usepackage{tabularx}
\usepackage{makecell}
\usepackage{multirow}
\usepackage{array}
\usepackage{graphicx}
\usepackage{colortbl}
\usepackage{caption}    
\usepackage{wrapfig}
\usepackage{adjustbox}
\usepackage{float}

\newcommand\eg{\textit{e.g.}}

\definecolor{sh_gray}{rgb}{0.84,0.84,0.84}
\definecolor{sh_gray2}{rgb}{1,0.89,0.75}
\definecolor{color3}{rgb}{0.95,0.95,0.95}
\definecolor{color4}{rgb}{0.96,0.96,0.86}
\definecolor{color5}{rgb}{0.90,0.90,0.90}

\usepackage[capitalize,noabbrev]{cleveref}
\theoremstyle{plain}

\theoremstyle{definition}

\theoremstyle{remark}

\usepackage[textsize=tiny]{todonotes}

\icmltitlerunning{LCUDiff: Latent Capacity Upgrade Diffusion}

\begin{document}

\twocolumn[
  \icmltitle{LCUDiff: Latent Capacity Upgrade Diffusion for\\ Faithful Human Body Restoration}

  \icmlsetsymbol{equal}{*}

\begin{icmlauthorlist}
  \icmlauthor{Jue Gong}{equal,sjtu}
  \icmlauthor{Zihan Zhou}{equal,sjtu}
  \icmlauthor{Jingkai Wang}{sjtu}
  \icmlauthor{Shu Li}{transsion}
  \icmlauthor{Libo Liu}{transsion}
  \icmlauthor{Jianliang Lan}{transsion}
  \icmlauthor{Yulun Zhang}{sjtu}
\end{icmlauthorlist}

\icmlaffiliation{sjtu}{Shanghai Jiao Tong University}
\icmlaffiliation{transsion}{Shenzhen Transsion Holdings Co., Ltd.}

\icmlcorrespondingauthor{Yulun Zhang}{yulun100@gmail.com}

  \vskip 0.3in
]

\printAffiliationsAndNotice{\icmlEqualContribution}  

\begin{abstract}
Existing methods for restoring degraded human-centric images often struggle with insufficient fidelity, particularly in human body restoration (HBR). Recent diffusion-based restoration methods commonly adapt pre-trained text-to-image diffusion models, where the variational autoencoder (VAE) can significantly bottleneck restoration fidelity. We propose LCUDiff, a stable one-step framework that upgrades a pre-trained latent diffusion model from the 4-channel latent space to the 16-channel latent space. For VAE fine-tuning, channel splitting distillation (CSD) is used to keep the first four channels aligned with pre-trained priors while allocating the additional channels to effectively encode high-frequency details. We further design prior-preserving adaptation (PPA) to smoothly bridge the mismatch between 4-channel diffusion backbones and the higher-dimensional 16-channel latent. In addition, we propose a decoder router (DeR) for per-sample decoder routing using restoration-quality score annotations, which improves visual quality across diverse conditions. Experiments on synthetic and real-world datasets show competitive results with higher fidelity and fewer artifacts under mild degradations, while preserving one-step efficiency. The code and model will be at \url{https://github.com/gobunu/LCUDiff}. 
\end{abstract}

\vspace{-3mm}
\section{Introduction}
Human body restoration (HBR) aims to restore high-quality (HQ) images of humans from low-quality (LQ) inputs. Human subjects are highly salient, making HBR especially sensitive to subtle identity and anatomical inconsistencies. In practical scenarios, images are often degraded during acquisition or transmission. Common degradations include motion blur from human movement, sensor noise, reduced spatial resolution, and artifacts introduced by JPEG compression. Restoration methods, therefore, seek high fidelity at both the pixel level and the perceptual level. This requirement is particularly critical for human-centric images. Insufficient fidelity can alter the subject’s appearance or produce anatomically implausible body parts. Such distortions severely limit the usefulness of restored human images in downstream applications, including face recognition~\cite{deng2019arcface}, keypoint detection~\cite{xu2022vitpose}, and 3D reconstruction~\cite{wang20213Dreview}.

\begin{figure}[t]
\begin{center}
\includegraphics[width=1.0\columnwidth]{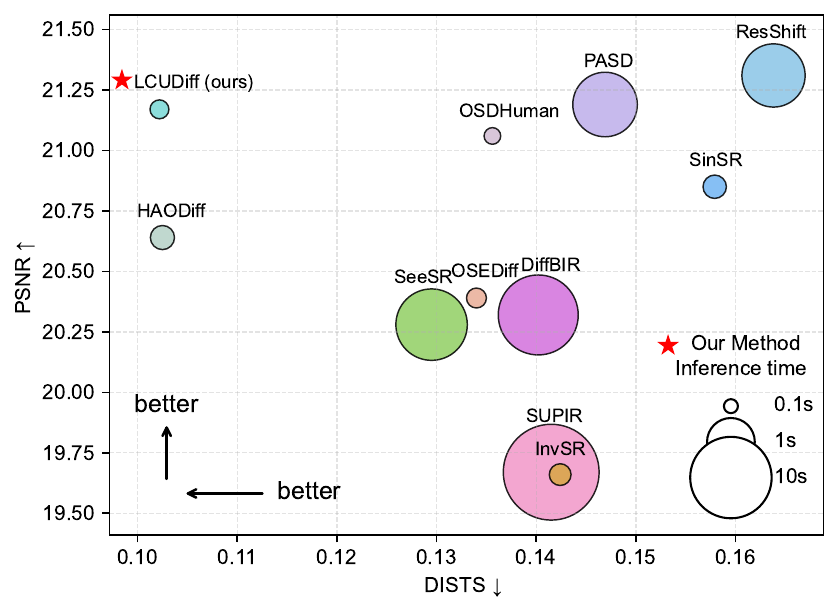}
\end{center}
\vspace{-4mm}
\caption{Trade-off between pixel-level and perceptual metrics on PERSONA-Val. Each point denotes a method, with DISTS on the x-axis and PSNR on the y-axis. The red star indicates our method. Inference time is measured on 512$\times$512 inputs using an NVIDIA RTX A6000. Most methods lie along a diagonal trend, suggesting that improving PSNR often comes at the cost of worse perceptual quality, or vice versa. Our method is located in the upper-left region, suggesting a better PSNR–DISTS balance.}
\label{fig:method_iqa}
\vspace{-7mm}
\end{figure}

In blind image restoration (BIR), a common paradigm is to adapt a pre-trained text-to-image (T2I) diffusion model, as its generative prior often yields more visually pleasing results~\cite{wu2024osediff,2024s3diff,wang2024osdface,yang2025sodiff}. Most modern approaches adopt latent diffusion models (LDM), where a separately pre-trained variational autoencoder (VAE) maps images into a compact latent space. Accordingly, the diffusion backbone is tightly coupled to that latent distribution and scaling. Increasing latent capacity typically relies on a higher-capacity autoencoder, \eg, by using more latent channels or a lower compression ratio. Although this can preserve finer structures, it also makes denoising more challenging. In practice, fully exploiting the added information often requires a larger denoiser, which increases computation and memory. Consequently, under efficiency-oriented settings such as limited denoising steps or lightweight backbones, an overly compressed autoencoder can become a fidelity bottleneck.

To improve fidelity under efficiency constraints, a common direction is to better exploit information in the LQ. Initializing LDM from the LQ latent and using LQ-derived guidance features can improve content consistency~\cite{wang2024osdface,gong2025osdhuman}. Another line of work increases latent capacity by upgrading the VAE~\cite{yi2025tvt,wang2025gendr}. These directions are complementary, improving faithfulness and raising the representational ceiling. However, pretrained diffusion backbones are tightly coupled to their original latent spaces, so unconstrained VAE changes can cause latent misalignment and artifacts.

\begin{figure}[t]
\scriptsize
\begin{center}

    \hspace{-5mm}
    \begin{adjustbox}{valign=t}
    \begin{tabular}{ccc}
    \includegraphics[width=0.135\textwidth]{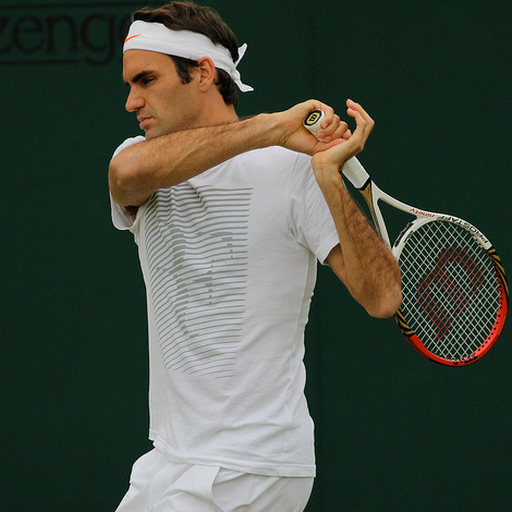} \hspace{-3.5mm} &
    \includegraphics[width=0.135\textwidth]{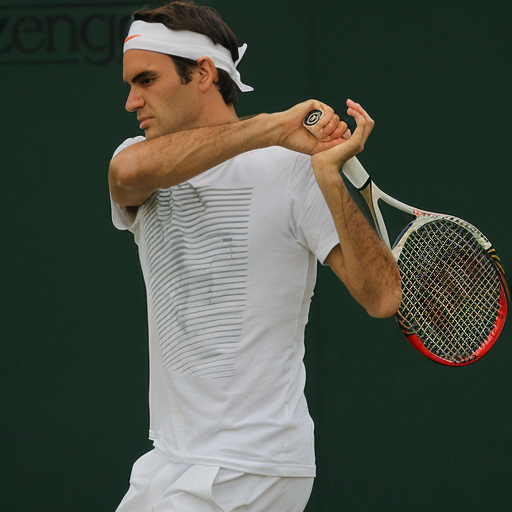} \hspace{-3.5mm} &
    \includegraphics[width=0.135\textwidth]{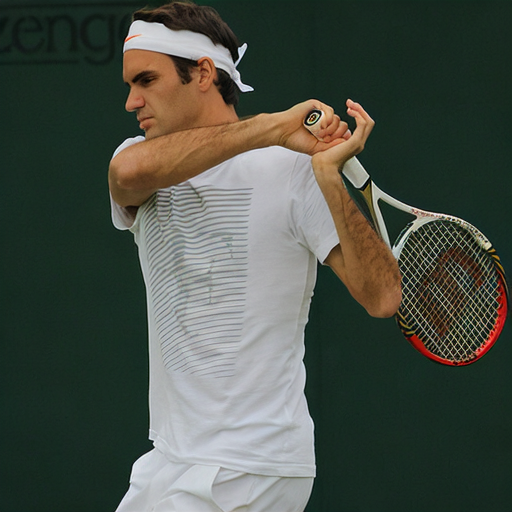} \hspace{-3.5mm} 
    \\
    
    Input\hspace{-3.5mm} &SD VAE\hspace{-3.5mm} &Our VAE\hspace{-3.5mm} \\
    \end{tabular}
    \end{adjustbox}
\end{center}

\vspace{-4.5mm}
\caption{Visual comparison of VAEs. Our VAE preserves subtle structures and reduces distortions compared with the SD VAE.}
\label{fig:comp_vae}
\vspace{-7mm}
\end{figure}

To address these limitations, we propose LCUDiff, a high-fidelity one-step diffusion framework for HBR. We expand the latent space from 4 to 16 channels and introduce channel splitting distillation (CSD), which keeps the first four anchor channels aligned with the pretrained latent space while using the remaining channels to carry high-frequency details, as illustrated in Fig.~\ref{fig:comp_vae}. To bridge the mismatch between 16-channel latents and the pretrained diffusion U-Net, we design prior-preserving adaptation (PPA) with dual input branches and a fusion schedule that smoothly transitions from the frozen prior pathway to the higher-dimensional latent pathway. These designs improve pixel-level and perceptual fidelity without increasing inference cost, and can be combined with existing conditional guidance for robustness. We further propose a decoder router (DeR) to select between the original and fine-tuned decoders per sample, leveraging their complementary behaviors to improve visual quality. Overall, as shown in Fig.~\ref{fig:method_iqa}, LCUDiff achieves strong fidelity and effective degradation removal.

In summary, we make the following four key contributions:
\vspace{-2mm}
\begin{itemize}
    \item We propose a 16-channel LCUDiff, and introduce channel splitting distillation (CSD) to substantially increase latent information capacity and detail expressiveness.
    \item We propose a training framework for channel expansion with prior-preserving adaptation (PPA) to bridge 16-channel latents and a 4-channel UNet, improving training stability while keeping inference efficient.
    \item We propose a decoder router (DeR) to select paths per sample based on degradation severity. This improves robustness under mild and extreme degradations.
    \item We deliver strong fidelity with competitive visual results under synthetic and real-world degradations, while preserving the efficient one-step inference.
\end{itemize}

\section{Related Work}
\label{sec:related_work}
\vspace{-2mm}

\subsection{Human Body Restoration}
\vspace{-1.5mm}
Human body restoration (HBR) is a critical area within blind image restoration (BIR).  It aims to recover geometric structures and textures with high fidelity from degraded inputs. Early research in HBR predominantly adapted general restoration paradigms based on generative adversarial networks (GANs)~\cite{goodfellow2014gan}. Researchers adopted this strategy to leverage the proven capabilities of these networks in texture synthesis. For instance, Real-ESRGAN~\cite{wang2021real} established a baseline using adversarial training for texture enhancement, while GFPGAN~\cite{wang2021gfpgan} was frequently incorporated as a cascade module to refine details in facial regions.

Recently, diffusion models have emerged as a powerful alternative for HBR due to their strong generative capacity. DiffBody~\cite{Zhang2024DiffBody} leverages generative diffusion priors to synthesize intricate details of the human body. Furthermore, HAODiff~\cite{gong2025haodiff} investigate diffusion frameworks operating in a single step, utilizing priors from models trained on a large scale to enable efficient inference and high-quality restoration from low-quality inputs.

\vspace{-2mm}
\subsection{Diffusion Models}
\vspace{-1mm}
Pre-trained text-to-image (T2I) models, especially Stable Diffusion (SD)~\cite{Rombach2022LDM}, have become a common backbone for generative image restoration. Multi-step methods such as PASD~\cite{yang2023pasd} and DiffBIR~\cite{lin2024diffbir} introduce control modules or degradation-aware encoders to guide iterative sampling. SeeSR~\cite{wu2024seesr} enhances content consistency with semantic priors, while SUPIR~\cite{yu2024supir} leverages larger foundation models to  extend restoration capability.

Research efforts also aim to accelerate diffusion inference. One-step frameworks such as OSEDiff~\cite{wu2024osediff} and SinSR~\cite{wang2024sinsr} typically rely on distillation-based training to enable fast generation. ResShift~\cite{yue2023resshift} targets efficient multi-step sampling by constructing a residual-shift diffusion trajectory. Latent Consistency Models (LCMs)~\cite{luo2023latent} employ consistency distillation to achieve high-fidelity synthesis with only a few steps. Notably, most of these approaches still operate in the standard VAE-compressed latent space with $f8$ downsampling and a fixed 4-channel representation~\cite{kingma2014vae,Rombach2022LDM}.

\begin{figure*}[t]
\begin{center}
\includegraphics[width=0.9\textwidth]{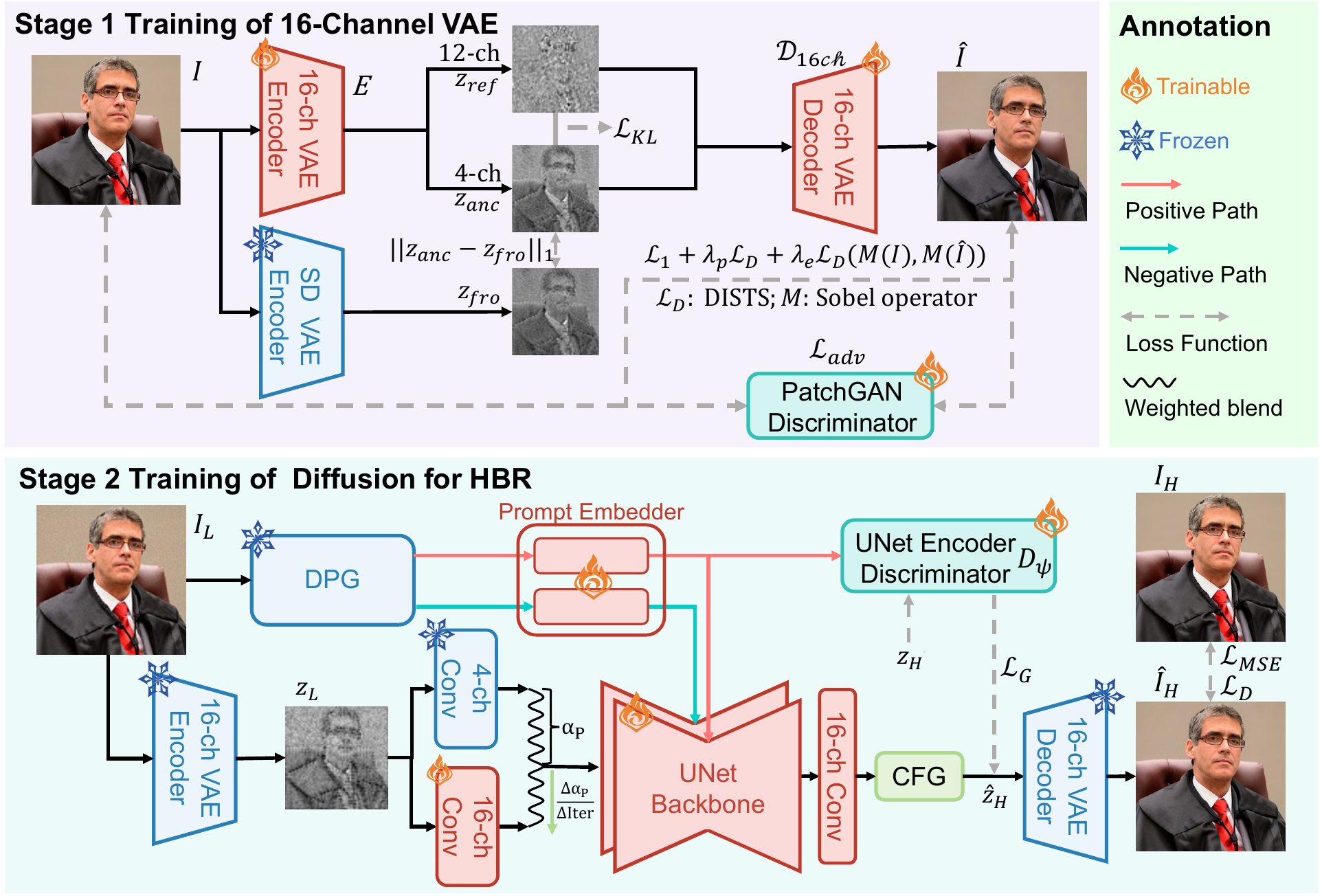}
\end{center}
\vspace{-4mm}
\caption{Model structure and training pipeline of our LCUDiff. \textbf{Stage 1}: We fine-tune a 16-channel VAE with channel splitting distillation (CSD). The first four anchor channels are aligned with the pretrained 4-channel latent space to preserve prior stability, while the remaining channels are optimized to encode additional high-frequency details. \textbf{Stage 2}: We train a one-step diffusion restoration model on the upgraded 16-channel latent space with prior-preserving adaptation (PPA). PPA builds two parallel input paths, an anchor-prior branch and a new 16-channel branch, and uses a fusion schedule to smoothly transition from the frozen prior pathway to the higher-dimensional latent pathway, stabilizing training without increasing inference overhead.}
\label{fig:model}
\vspace{-7mm}
\end{figure*}

\subsection{Latent Representation and Autoencoders}
\vspace{-1mm}
Latent diffusion models (LDM)~\cite{Rombach2022LDM} use a VAE to compress images into a low-resolution latent space. In Stable Diffusion (e.g., SD 2.1), the standard VAE adopts an $f8$ downsampling factor with a fixed 4-channel latent. Recent foundation models scale up VAE capacity and increase latent channels to learn richer representations and retain more information under compression.

Recent work modifies VAEs or latent representations to better support generation and restoration, often trading off reconstruction fidelity and efficiency. TVT~\cite{yi2025tvt} substantially improves fidelity by reducing the downsampling factor to $f4$, preserving more spatial information in the latent space. InvSR~\cite{yue2024invsr} pursues higher information retention for $\times$4 super-resolution by explicitly redesigning the encoder as a noise predictor and using only $\times$2 downsampling relative to the low-resolution input, avoiding the heavy information loss of the standard $f8$ setting. In contrast, FastVSR~\cite{li2025fastvsr} emphasizes efficiency for video super-resolution with an asymmetric autoencoder, combining a standard encoder and a highly compressed decoder with indirect upsampling.

\section{Method}

\subsection{16-Channel Variational Autoencoder}
\label{sec:vae}

Standard latent diffusion models (LDM)~\cite{podell2023sdxl} typically use a variational autoencoder (VAE)~\cite{kingma2014vae} with KL regularization. This VAE has an $8\times$ downsampling factor and a fixed 4-channel latent space. This setup balances efficiency and generation quality. However, its high compression ratio creates a severe bottleneck for human body restoration (HBR). Critical details at high frequencies, such as clothing textures and hair, are often lost. This limits the upper bound of restoration fidelity.

We analyze the SD-Turbo~\cite{stabilityai_sdturbo_2023} VAE as a representative baseline. Its backbone network outputs 512 feature channels. However, the subsequent convolution layer compresses these into only 8 channels that represent the mean and the variance. This results in a final latent code of just 4 channels. This drastic reduction from 512 channels to 4 channels causes significant information loss. It is a key contributor to missing details at high frequencies. To address this, we propose to fine-tune a 16-Channel Variational Autoencoder. By expanding the latent space, we aim to improve information retention. However, the downstream diffusion UNet expects the original 4-channel distribution. A VAE of high dimension can introduce a distribution mismatch with the pre-trained diffusion priors.

\textbf{Channel Splitting Distillation.} 
To effectively bridge this distributional gap, we introduce a specialized training strategy. Previous studies, such as Lightning-DiT ~\cite{yao2025vavae} and DCAE1.5 ~\cite{chen2025DCAE15}, suggest that VAEs configured with high channel capacity require a highly structured latent space. Specifically, they indicate that information at low frequencies must remain dominant to ensure generative stability. Guided by this critical insight, we propose the channel splitting distillation (CSD) strategy.

We explicitly partition the 16 latent channels into two groups: anchor channels $z_{anc}$ and refine channels $z_{ref}$. The $z_{anc}$ group consists of the first 4 channels, which serve as the semantic foundation. During training, we apply an $\mathcal{L}_1$ distillation loss to $z_{anc}$. This forces it to closely align with the output of the frozen pre-trained SD VAE. This constraint ensures compatibility with the diffusion priors. Meanwhile, $z_{ref}$, which comprises the remaining 12 channels,  captures high-frequency residuals. These are the subtle details that the anchor channels fail to encode.

\textbf{Training Objective.} 
Our objective function consists of three distinct components, namely reconstruction, regularization, and adversarial constraints.

We combine the loss $\mathcal{L}_1$ with the perceptual distance DISTS~\cite{ding2020dists} $\mathcal{L}_\text{D}$ as the reconstruction term to preserve the fidelity of the reconstruction. In addition, we introduce an explicit edge-aware enhancement to increase the sensitivity of DISTS to boundary structures. Specifically, we compute the spatial image gradients $G_x$ and $G_y$ using the standard Sobel kernels $S_x$ and $S_y$:
\begin{equation}
    S_x = \begin{bmatrix} 1 & 0 & -1 \\ 2 & 0 & -2 \\ 1 & 0 & -1 \end{bmatrix}, \quad 
    S_y = \begin{bmatrix} 1 & 2 & 1 \\ 0 & 0 & 0 \\ -1 & -2 & -1 \end{bmatrix}.
\end{equation}
The resulting edge magnitude map is defined as $M(I) = \sqrt{G_x^2 + G_y^2}$. Consequently, the total reconstruction objective $\mathcal{L}_\text{rec}$ is formulated as follows:
\begin{equation}
    \mathcal{L}_\text{rec} \!= \! \mathcal{L}_1(I,\hat{I}) \!+\! \lambda_{p} \mathcal{L}_\text{D}(I, \hat{I})\!+ \!\lambda_{e} \mathcal{L}_\text{D}\left(M(I),M(\hat{I})\right).
\end{equation}
\label{eq:vae_rec}
The regularization component serves to strictly enforce the proposed channel alignment strategy. It includes the standard KL divergence loss combined with the explicit CSD loss for the anchor channels $z_{anc}$:
\begin{equation}
    \mathcal{L}_\text{reg} = \lambda_\text{KL} \mathcal{L}_\text{KL}(z) + \lambda_\text{CSD}\mathcal{L}_1(z_{anc},z_{fro}),
\end{equation}
where $z_{fro}$ denotes the latent code representation that is derived from the fixed pre-trained SD VAE.

To improve perceptual realism, we use a patch-based discriminator loss $\mathcal{L}_\text{adv}$, following the standard setup of LDM~\cite{podell2023sdxl}. The total training objective $\mathcal{L}_\text{VAE}$ is the weighted sum of these parts:
\begin{equation}
    \mathcal{L}_\text{VAE} = \mathcal{L}_\text{rec} + \mathcal{L}_\text{reg} + \lambda_\text{adv} \mathcal{L}_\text{adv}.
\end{equation}

\subsection{Diffusion for Human Body Restoration}
\label{sec:diffusion}

Standard text prompts cannot capture pixel-level degradation details. Therefore, we employ a dual-prompt guidance (DPG)~\cite{gong2025haodiff} module. This module extracts two distinct feature sets from the low-quality input: positive cues ($\mathbf{F}_{pos}$) and negative cues ($\mathbf{F}_{neg}$). The positive cues explicitly represent the reconstruction targets, while the negative cues model degradation artifacts such as blur and noise. However, the dimensions and feature distributions of these cues differ significantly from the standard text embeddings used by the frozen UNet. To inject the dual visual prompts into the text-conditioned UNet, we employ a lightweight feature embedder. First, a downsampling module utilizes two consecutive convolutional layers with a stride of 2. This achieves a $4\times$ spatial compression and effectively reduces the sequence length. The compressed sequence is then processed by two Transformer encoder layers~\cite{vaswani2017attention} followed by attention pooling~\cite{lee2019attentionpool}. This process yields the final embeddings $\mathbf{p}_{pos}$ and $\mathbf{p}_{neg}$, which are compatible with the SD text embedding space.

We use the generative prior of pre-trained LDM for high-fidelity HBR. Given an LQ input $I_L$, we first encode it into the latent space by a VAE encoder $E$: $z_L = E(I_L)$. Instead of starting from pure Gaussian noise, we directly treat $z_L$ as a degraded latent and perform a single denoising step conditioned on a timestep $\tau$. Following LDM, the forward noise schedule is parameterized by $\bar{\alpha}_t=\prod_{s=1}^{t}(1-\beta_s)$. Given a latent $z_\tau$ and the predicted noise $\varepsilon_{noise}$, the clean latent estimate is formally defined as:
\begin{equation}
\hat{z}_0=\frac{z_\tau - \sqrt{1 - \bar{\alpha}_\tau}\, \varepsilon_{noise}}{\sqrt{\bar{\alpha}_\tau}}.
\end{equation}
We set $z_\tau=z_L$ and output the noise-free latent $\hat{z}_H \equiv \hat{z}_0$. We empirically fix $\tau=249$ for all experiments.

Subsequently, we incorporate classifier-free guidance (CFG)~\cite{ho2021cfg} during training. Let $\varepsilon_{pos}$ and $\varepsilon_{neg}$ denote the noise predictions conditioned on the positive and negative embeddings ($\mathbf{p}_{pos}$ and $\mathbf{p}_{neg}$), respectively. We form the guided noise prediction as $\varepsilon_\text{noise} = \varepsilon_{neg} + \lambda_{cfg}\,(\varepsilon_{pos} - \varepsilon_{neg})$, and use $\varepsilon_\text{noise}$ to steer the model toward the high-quality prior. To preserve the generative capability while keeping training efficient, we fine-tune the UNet backbone with LoRA~\cite{hu2022lora}.

\textbf{Prior-Preserving Adaptation.}
To bridge the channel mismatch, we introduce a lightweight prior-preserving adaptation (PPA) strategy. It reconciles our fine-tuned 16-channel VAE with the standard 4-channel UNet. For the input, we use a blended-convolution design with two parallel branches. The prior branch keeps only the first four anchor channels $z_{anc}$ and feeds them into the frozen original input layer. The new branch learns a direct pathway from the full 16-channel latent to the UNet backbone. The two branches are fused using an iteration-dependent training scalar $\alpha_{\mathrm{P}}$. During early training, $\alpha_{\mathrm{P}}$ linearly transitions from using only $z_{anc}$ to using the full 16-channel latent. This schedule gradually shifts the model from the frozen prior to the learned 16-channel pathway and improves training stability. On the output side, we expand the prediction head to produce 16 channels. We initialize the first four channels with pre-trained weights to retain the original prior at the beginning of training.

\textbf{Training Objective.}
Our loss functions supervise the synthesized latent $\hat{z}_H$ and the final image. First, we adopt the mean squared error (MSE) loss $\mathcal{L}_{\text{MSE}}$ to minimize reconstruction errors. To better match human visual preference, we incorporate DISTS $\mathcal{L}_\text{D}$, a human-aligned image distance, as an auxiliary loss. Furthermore, following previous work~\cite{li2025d3sr}, use the pre-trained SD UNet itself as a discriminator $D_\psi$ to effectively address latent distribution distortions. This adversarial component ensures that the guided $\hat{z}_H$ aligns closely with the realistic high-quality data distribution. The total loss function $\mathcal{L}_\text{total}$ is:
\begin{equation}
    \mathcal{L}_\text{total} = \mathcal{L}_\text{MSE}(\hat{I}_H, I_H) + \mathcal{L}_\text{D}(\hat{I}_H, I_H) + \lambda_\text{G} \mathcal{L}_\text{G}(\hat{z}_H).
\end{equation}
\label{eq:model_loss}
The objective for the generator and discriminator comprises the generator loss $\mathcal{L}_\text{G}$ and the discriminator loss $\mathcal{L}_\text{Dis}$:
\begin{equation}
\begin{split}
    \mathcal{L}_{G}(\hat{z}_H) = & -\mathbb{E}_t [\log D_\psi (\mathcal{F}(\hat{z}_H, t))], \\
    \mathcal{L}_\text{Dis}(\hat{z}_H, z_H) = & -\mathbb{E}_t [\log (1 - D_\psi (\mathcal{F}(\hat{z}_H, t)))] \\
     & - \mathbb{E}_t [\log D_\psi (\mathcal{F}(z_H, t))],
\end{split}
\end{equation}
where $\mathcal{F}(\cdot)$ represents the diffusion noise addition process.

\begin{figure}[t]
\begin{center}
\includegraphics[width=1.0\columnwidth]{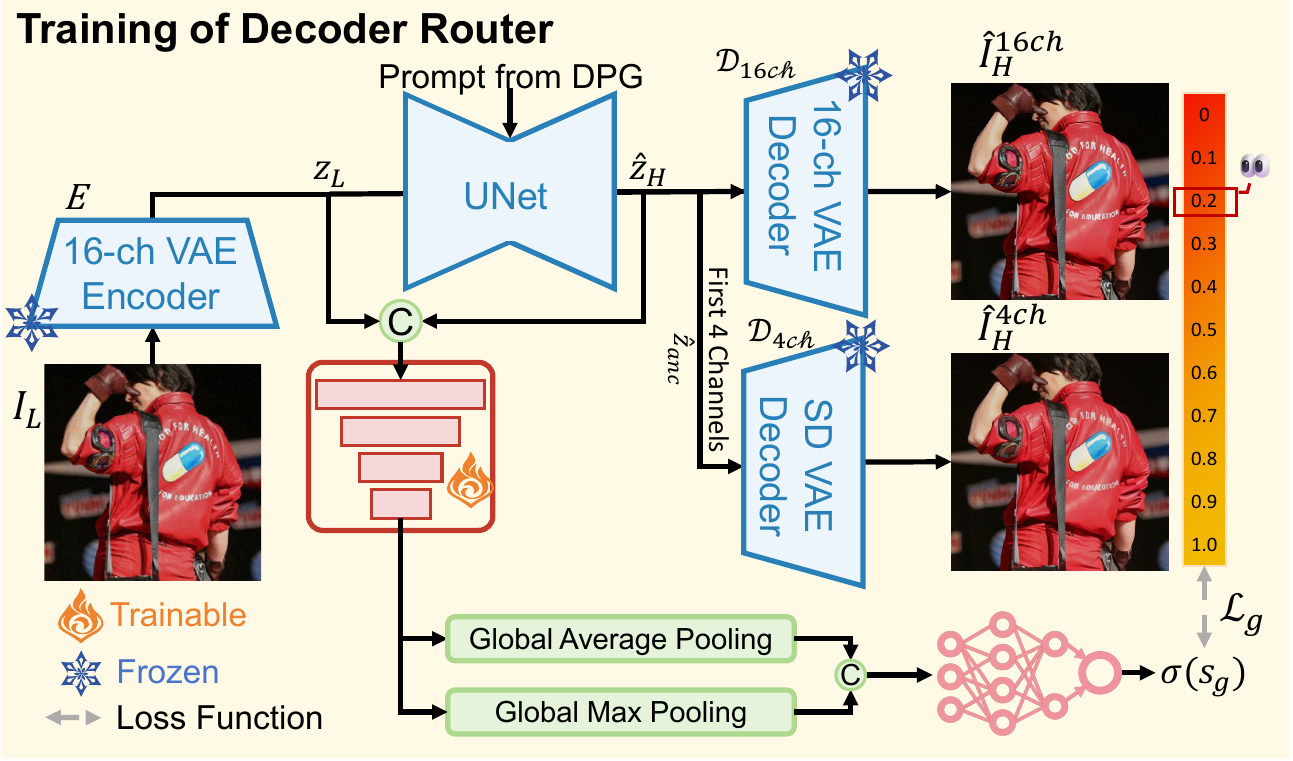}
\end{center}
\vspace{-4mm}
\caption{Training of decoder router (DeR). We build a preference dataset by decoding each restored latent with both the pretrained $\mathcal{D}_{4ch}$ and the fine-tuned $\mathcal{D}_{16ch}$. DeR takes the concatenation of $z_L$ and $\hat{z}_H$ as input and is trained with a soft BCE loss to predict the decoder preference. Here $\sigma(\cdot)$ denotes the sigmoid function, and the diffusion backbone and both decoders are kept frozen.}
\label{fig:decoder_router}
\vspace{-6mm}
\end{figure}

\subsection{Adaptive Decoding via Decoder Router (DeR)}
\label{sec:decoderrouter}
Following the distillation training based on channel splitting described in Sec.~\ref{sec:vae}, we observe a unique dual compatibility in the restored latent $\hat{z}_H$. Specifically, the Anchor Channels ($\hat{z}_{anc}$) align with the standard SD VAE latent space. They can be decoded directly by the pre-trained decoder $\mathcal{D}_{4ch}$. In contrast, the Complete Latent ($\hat{z}_{H}$) corresponds to the decoder $\mathcal{D}_{16ch}$ obtained via fine-tuning.

Experiments show that these two pathways have complementary strengths. $\mathcal{D}_{4ch}$ relies on the stable generative prior. It provides reliable structural constraints, ideal for samples with extreme degradation. Conversely, $\mathcal{D}_{16ch}$ excels at recovering textures of high frequency. It ensures fuller detail preservation for samples with normal degradation. This observation motivates a routing mechanism for each sample instead of a fixed choice. We design DeR to assign each latent code to its optimal decoder adaptively.

\textbf{Network Architecture.}
We propose a lightweight convolutional gating network to predict the routing score. The network takes the concatenation of the low-quality latent $z_{L}$ and the restored latent $\hat{z}_H$ as input. This fuses degradation cues with the restoration results. The backbone architecture consists of four stacked convolution blocks. Each block sequentially applies a $3\times3$ convolution, group normalization~\cite{wu2018group}, and SiLU activation~\cite{Elfwing2018silu} twice. The final three blocks use a stride of 2 to expand the receptive field progressively.

For feature aggregation, we apply both global average pooling (GAP) and global max pooling (GMP). This dual-pooling strategy captures both holistic statistics and distinct discriminative anomalies. Finally, a classification head with dropout processes the aggregated features. It outputs a scalar logit $s_{g}$, representing the preference probability.

\textbf{Training and Inference.}
To effectively train DeR, we create a preference dataset using the fine-tuned diffusion model. For each sample, we generate outputs from both decoders. We then assign a label $y \in [0,1]$ in steps of 0.1 based on restoration quality, where $y=0$ favors the 16-channel output. Crucially, we optimize only the DeR module, keeping the diffusion backbone and decoders frozen. The objective is the soft binary cross entropy (BCE) loss:
\begin{equation}
    \mathcal{L}_{g} = - \big[ y \log(\sigma(s_{g}))  + (1 - y) \log(1 - \sigma(s_{g})) \big],
\end{equation}
where $\sigma(\cdot)$ is the sigmoid activation function. 

During the inference phase, we use a hard routing strategy for efficiency. If the predicted probability $\sigma(s_{g})$ exceeds the threshold, we use $\mathcal{D}_{4ch}$. Otherwise, we use $\mathcal{D}_{16ch}$. This ensures that we execute only one decoder per sample. The computational latency overhead is negligible.

\begin{table*}[h] 
\small
\setlength{\tabcolsep}{0.5mm}
\renewcommand{\arraystretch}{1.05}
\centering
\newcolumntype{C}{>{\centering\arraybackslash}X}
\begin{tabularx}{1\textwidth}{l|CCCCCC|CCCC}
\toprule
\rowcolor{color3} \textbf{Methods} 
& PSNR$\uparrow$ & PSNRY$\uparrow$ & SSIM$\uparrow$ & SSIMY$\uparrow$ & DISTS$\downarrow$ 
& LPIPS$\downarrow$ & C\mbox{-}IQA$\uparrow$ & H\mbox{-}IQA$\uparrow$ & TOPIQ$\uparrow$ & TRES$\uparrow$ \\
\midrule
SUPIR~\cite{yu2024supir} & 19.67 & 21.43 & 0.5451 & 0.5775 & 0.1415 & 0.2929 & \textbf{0.7908} & 0.7329 & 0.7464 & 93.1644 \\ 
DiffBIR~\cite{lin2024diffbir} & 20.32 & 22.07 & 0.5592 & 0.5946 & 0.1402 & 0.2797 & 0.7792 & \textbf{0.7616} & \textbf{0.7672} & \textbf{96.5071} \\
SeeSR~\cite{wu2024seesr} & 20.28 & 22.04 & 0.5902 & 0.6243 & 0.1295 & 0.2555 & 0.7620 & 0.7613 & 0.7747 & 98.8910 \\ 
PASD~\cite{yang2023pasd} & 21.19 & 22.93 & 0.6248 & 0.6542 & 0.1469 & 0.2910 & 0.6121 & 0.6732 & 0.6564 & 86.5425 \\ 
ResShift~\cite{yue2023resshift} & \textbf{21.31} & \textbf{23.16} & \textbf{0.6283} & \textbf{0.6591} & 0.1638 & 0.2848 & 0.5365 & 0.5501 & 0.5000 & 71.2869 \\ 
\midrule
SinSR~\cite{wang2024sinsr} & 20.85 & 22.70 & 0.6007 & 0.6317 & 0.1579 & 0.2840 & 0.6037 & 0.6192 & 0.5833 & 82.1876 \\ 
OSEDiff~\cite{wu2024osediff} & 20.39 & 22.13 & 0.6049 & 0.6391 & 0.1340 & 0.2507 & 0.6961 & 0.6641 & 0.6376 & 86.8851 \\ 
InvSR~\cite{yue2024invsr} & 19.66 & 21.44 & 0.5844 & 0.6165 & 0.1424 & 0.2709 & 0.6886 & 0.6593 & 0.6383 & 85.2479 \\ 
OSDHuman~\cite{gong2025osdhuman} & \textcolor{blue}{21.06} & \textcolor{blue}{22.79} & \textcolor{red}{0.6180} & \textcolor{red}{0.6505} & 0.1356 & 0.2384 & 0.7312 & 0.6205 & 0.5870 & 79.2763 \\ 
HAODiff~\cite{gong2025haodiff} & 20.64 & 22.32 & 0.6037 & 0.6331 & \textcolor{blue}{0.1025} & \textcolor{red}{\textbf{0.2047}} & \textcolor{red}{0.7728} & \textcolor{blue}{0.6656} & \textcolor{blue}{0.6712} & \textcolor{blue}{88.5220} \\
\midrule   
LCUDiff (ours) & \textcolor{red}{21.17} & \textcolor{red}{22.93} & \textcolor{blue}{0.6108} & \textcolor{blue}{0.6416} & \textcolor{red}{\textbf{0.1022}} & \textcolor{blue}{0.2149} & \textcolor{blue}{0.7540} & \textcolor{red}{0.6735} & \textcolor{red}{0.6715} & \textcolor{red}{89.3537} \\ 
\bottomrule
\end{tabularx}
\vspace{0.5mm}
\caption{Quantitative results on PERSONA-Val. The best and second-best results among one-step diffusion methods are highlighted in \textcolor{red}{red} and \textcolor{blue}{blue}, respectively. The best results among all methods are shown in \textbf{bold}. C-IQA stands for CLIPIQA, H-IQA stands for HYPERIQA. }
\vspace{-7mm}
\label{table:model_metrics}
\end{table*}

\section{Experiments}

\subsection{Experimental Settings}
\label{sec:setup}
\textbf{Training Datasets.}
For the fine-tuning stage of VAE, we construct a composite training set. This set comprises the PERSONA~\cite{gong2025osdhuman} dataset, the LSDIR~\cite{Li2023LSDIR} dataset, and a subset of 20k images sampled from FFHQ~\cite{karras2019ffhq}. Specifically, we randomly crop images from LSDIR to $512\times512$. The images from FFHQ are downsampled beforehand to the same resolution. During the training of the diffusion UNet, we reduce the LSDIR component to 20k samples. This adjustment biases the training distribution toward human body restoration. We synthesize paired HQ and LQ images using the degradation pipeline described in HAODiff~\cite{gong2025haodiff}. The training data of DeR is sampled from PERSONA 4k images and assigned preference labels. Detailed dataset settings are provided in the supplementary material.

\textbf{Testing Datasets.}
For comprehensive evaluation, we utilize the synthetic set PERSONA-Val~\cite{gong2025osdhuman} and the real-world set MPII-Test~\cite{gong2025haodiff}. The PERSONA-Val set is generated using the same degradation pipeline as the training stage to assess restoration performance under complex degradation. Additionally, we employ a separate light degradation pipeline to specifically evaluate the restoration fidelity of one-step diffusion models.

\textbf{Evaluation Metrics.} 
We employ a comprehensive set of metrics to evaluate restoration performance on PERSONA-Val. For full-reference fidelity assessment, we utilize PSNR, SSIM~\cite{wang2004ssim}, DISTS~\cite{ding2020dists}, and LPIPS~\cite{zhang2018lpips}. Furthermore, we calculate PSNR and SSIM specifically on the luminance Y channel of the YCbCr color space, denoted as PSNRY and SSIMY. For quality assessment with no reference, we adopt CLIPIQA~\cite{wang2022clipiqa}, HyperIQA~\cite{Su2020hyperiqa}, TOPIQ~\cite{Chen2024topiq}, and TReS~\cite{golestaneh2021tres}. These metrics with no reference are also applied to evaluate the real-world MPII-Test dataset.

\begin{figure*}[t]

\scriptsize
\begin{center}

\scalebox{0.98}{

    \hspace{-0.4cm}
    \begin{adjustbox}{valign=t}
    \begin{tabular}{ccccccccccc}
    \includegraphics[width=0.09\textwidth]{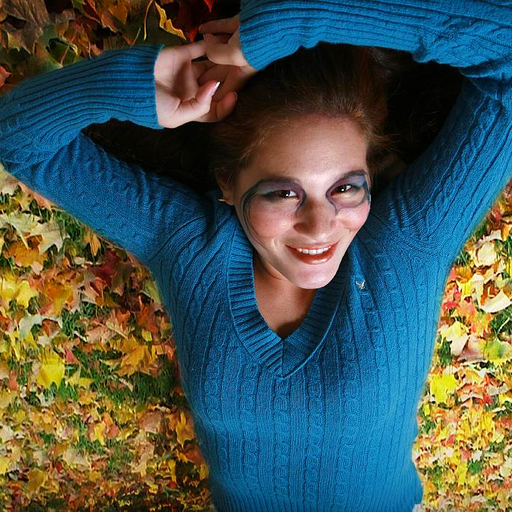} \hspace{-4.5mm} &
    \includegraphics[width=0.09\textwidth]{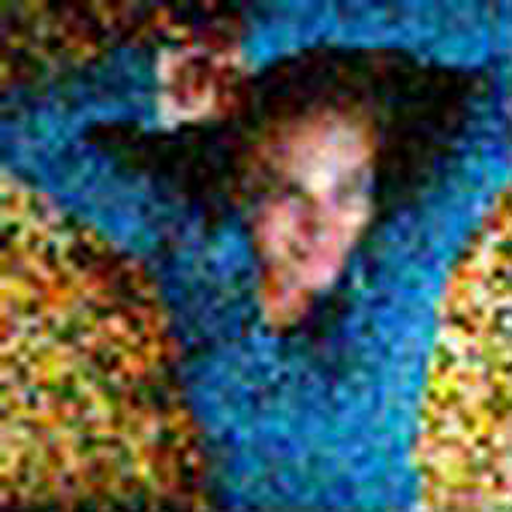} \hspace{-4.5mm} &
    \includegraphics[width=0.09\textwidth]{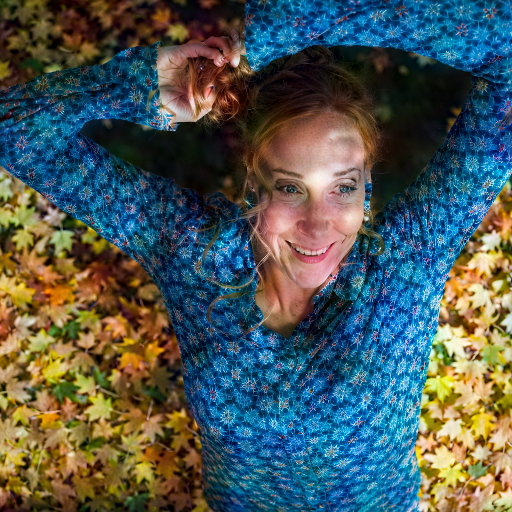} \hspace{-4.5mm} &
    \includegraphics[width=0.09\textwidth]{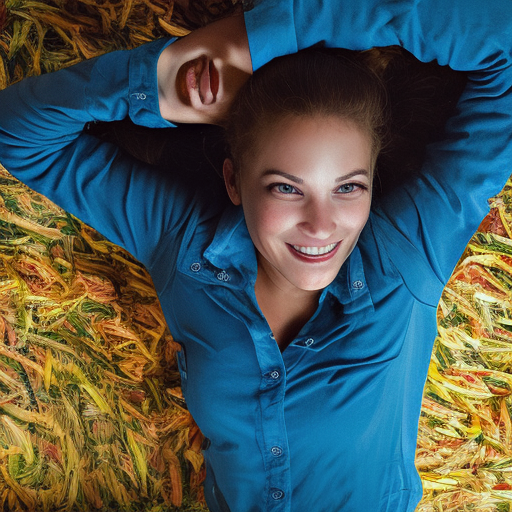} \hspace{-4.5mm} &
    \includegraphics[width=0.09\textwidth]{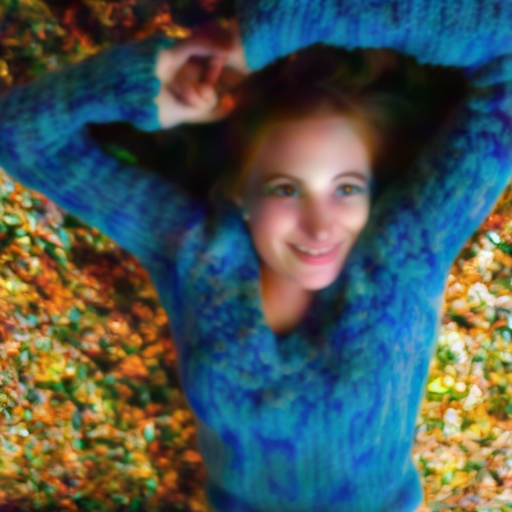} \hspace{-4.5mm} &
    \includegraphics[width=0.09\textwidth]{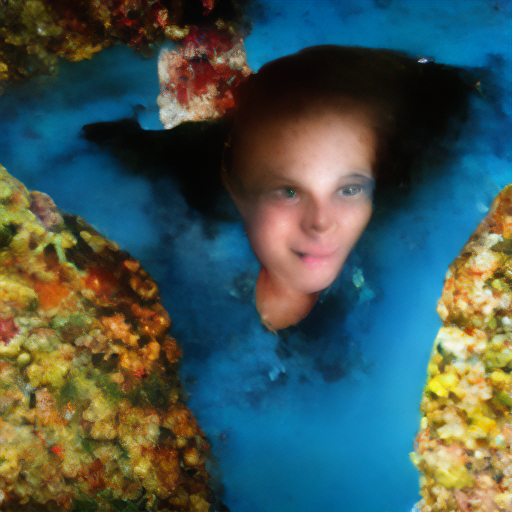} \hspace{-4.5mm} &
    \includegraphics[width=0.09\textwidth]{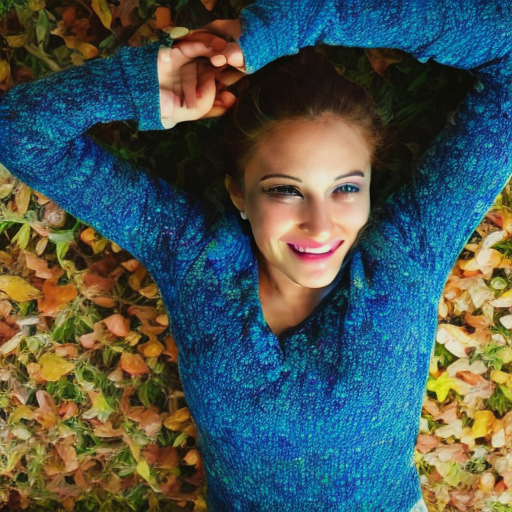} \hspace{-4.5mm} &
    \includegraphics[width=0.09\textwidth]{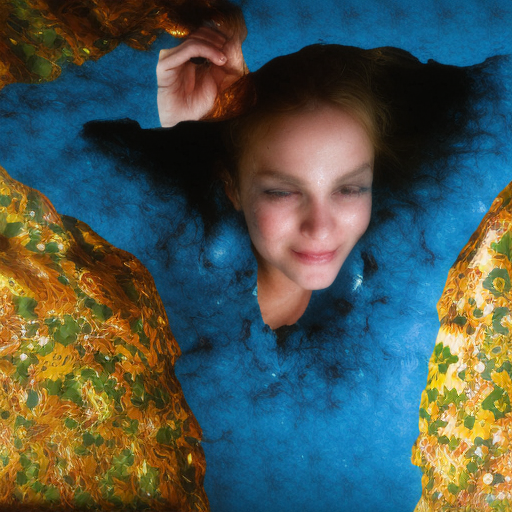} \hspace{-4.5mm} &
    \includegraphics[width=0.09\textwidth]{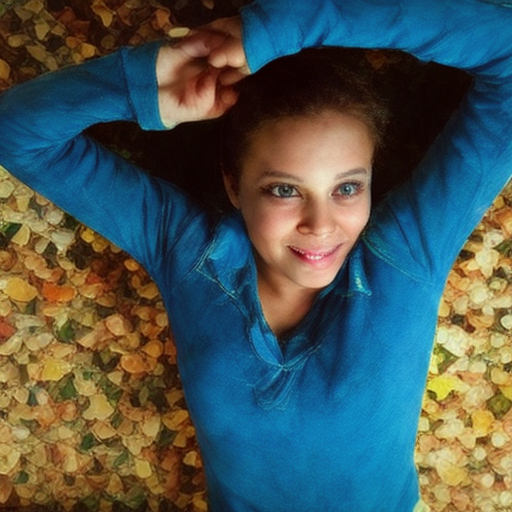} \hspace{-4.5mm} &
    \includegraphics[width=0.09\textwidth]{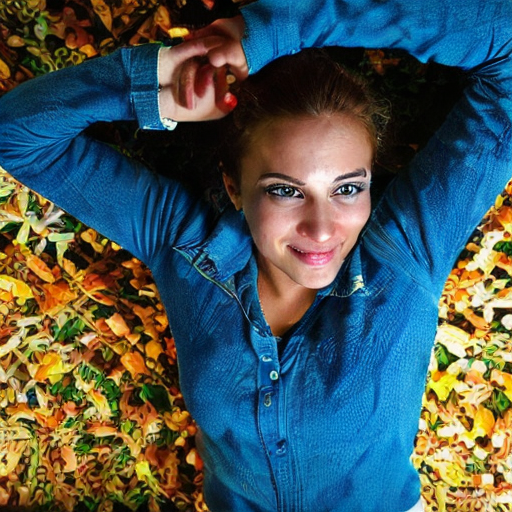} \hspace{-4.5mm} &
    \includegraphics[width=0.09\textwidth]{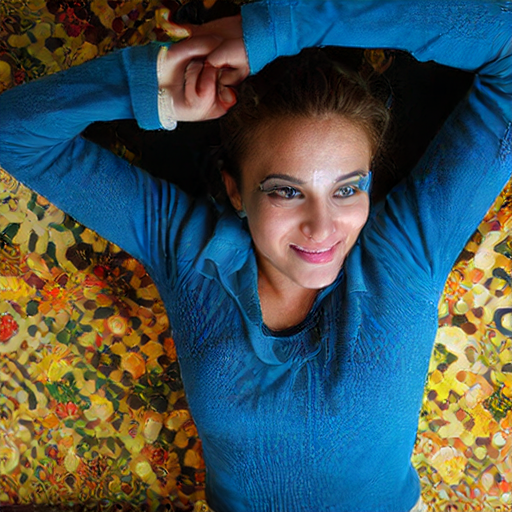} \hspace{-4.5mm} 
    \\
    \end{tabular}
    \end{adjustbox}
    
}
\scalebox{0.98}{

    \hspace{-0.4cm}
    \begin{adjustbox}{valign=t}
    \begin{tabular}{ccccccccccc}
    \includegraphics[width=0.09\textwidth]{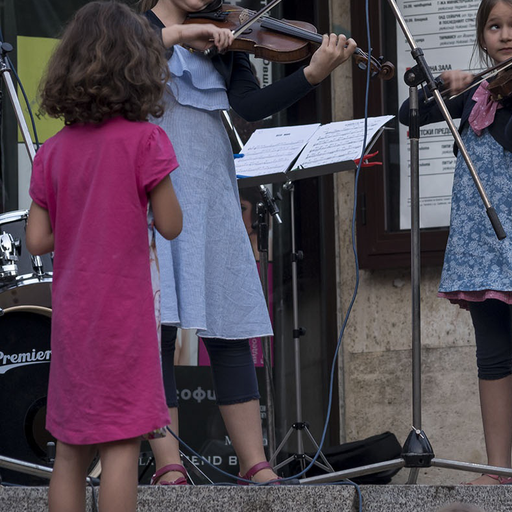} \hspace{-4.5mm} &
    \includegraphics[width=0.09\textwidth]{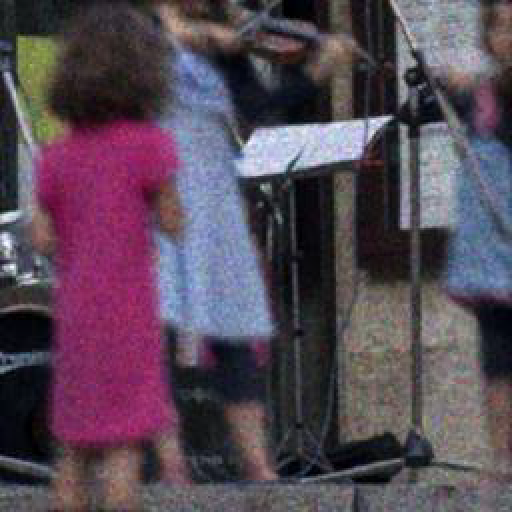} \hspace{-4.5mm} &
    \includegraphics[width=0.09\textwidth]{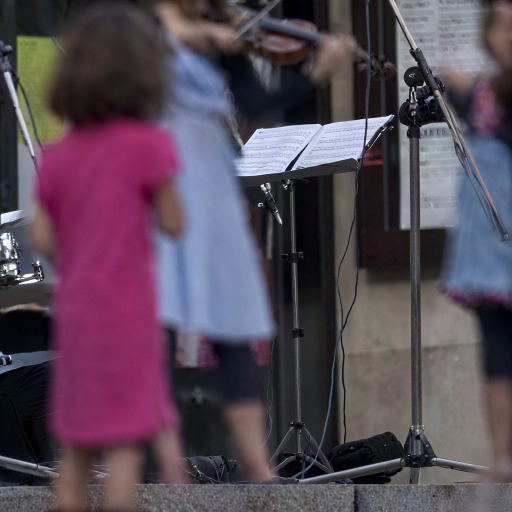} \hspace{-4.5mm} &
    \includegraphics[width=0.09\textwidth]{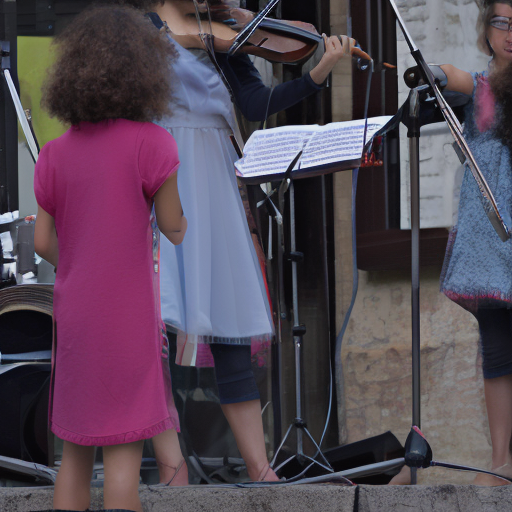} \hspace{-4.5mm} &
    \includegraphics[width=0.09\textwidth]{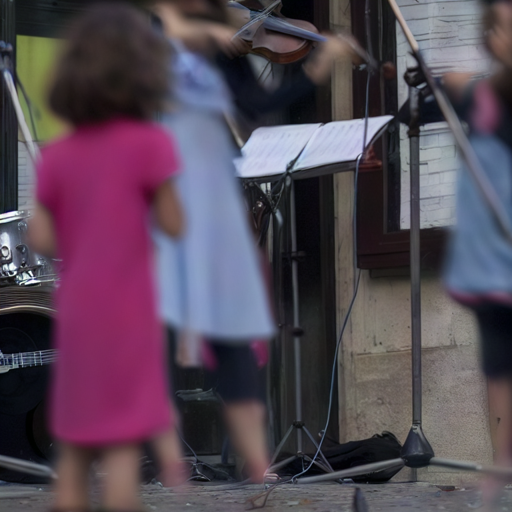} \hspace{-4.5mm} &
    \includegraphics[width=0.09\textwidth]{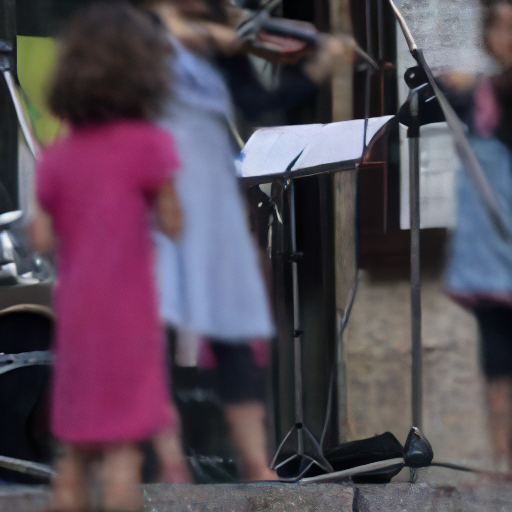} \hspace{-4.5mm} &
    \includegraphics[width=0.09\textwidth]{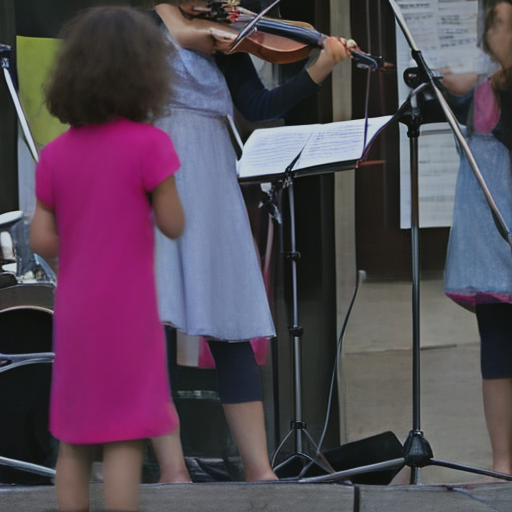} \hspace{-4.5mm} &
    \includegraphics[width=0.09\textwidth]{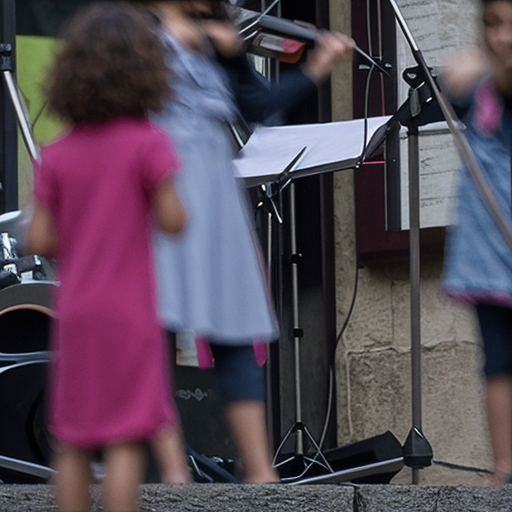} \hspace{-4.5mm} &
    \includegraphics[width=0.09\textwidth]{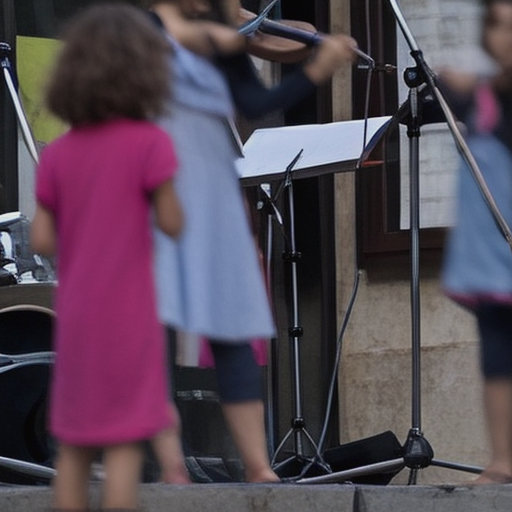} \hspace{-4.5mm} &
    \includegraphics[width=0.09\textwidth]{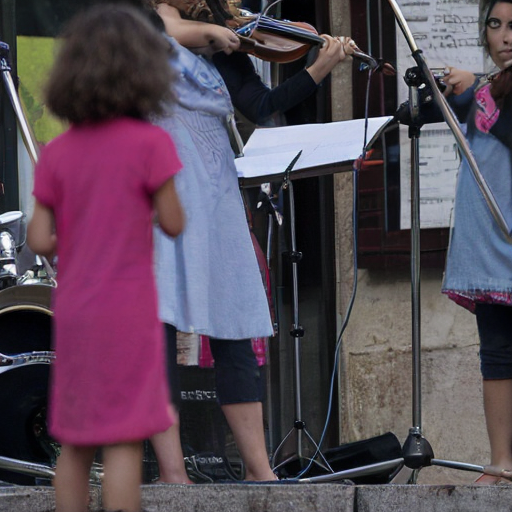} \hspace{-4.5mm} &
    \includegraphics[width=0.09\textwidth]{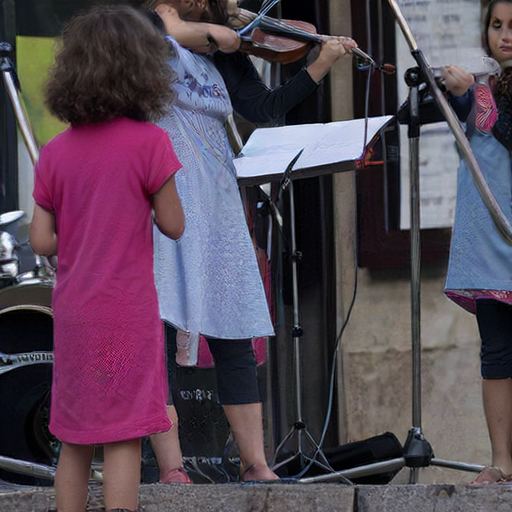} \hspace{-4.5mm} 
    \\
    \end{tabular}
    \end{adjustbox}
    
}
\scalebox{0.98}{

    \hspace{-0.4cm}
    \begin{adjustbox}{valign=t}
    \begin{tabular}{ccccccccccc}
    \includegraphics[width=0.09\textwidth]{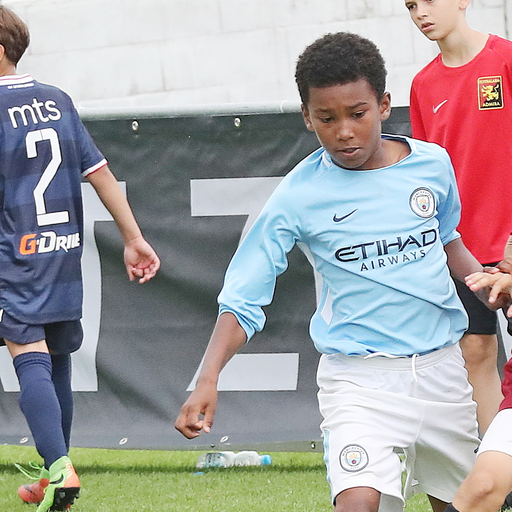} \hspace{-4.5mm} &
    \includegraphics[width=0.09\textwidth]{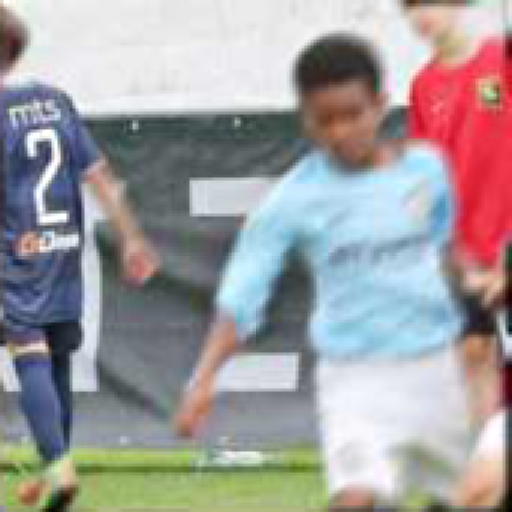} \hspace{-4.5mm} &
    \includegraphics[width=0.09\textwidth]{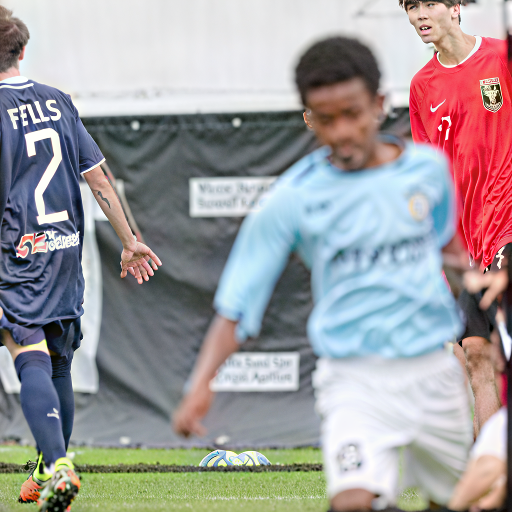} \hspace{-4.5mm} &
    \includegraphics[width=0.09\textwidth]{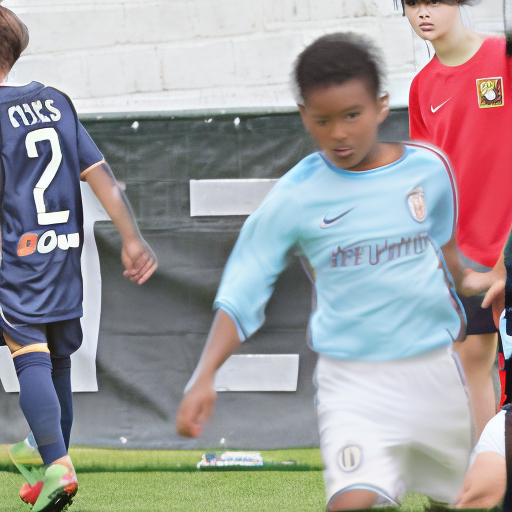} \hspace{-4.5mm} &
    \includegraphics[width=0.09\textwidth]{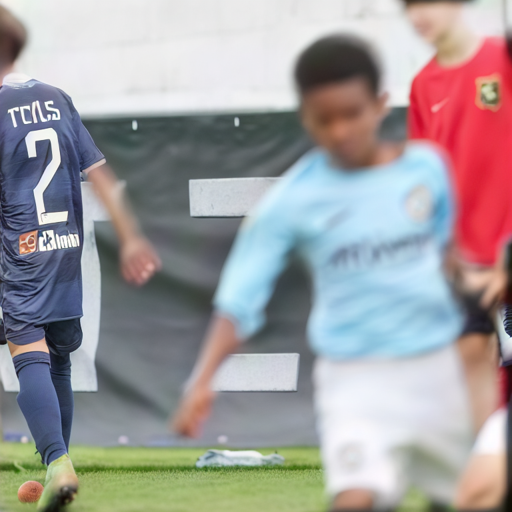} \hspace{-4.5mm} &
    \includegraphics[width=0.09\textwidth]{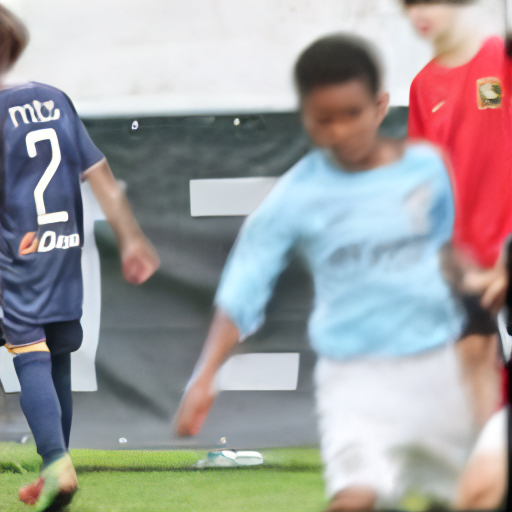} \hspace{-4.5mm} &
    \includegraphics[width=0.09\textwidth]{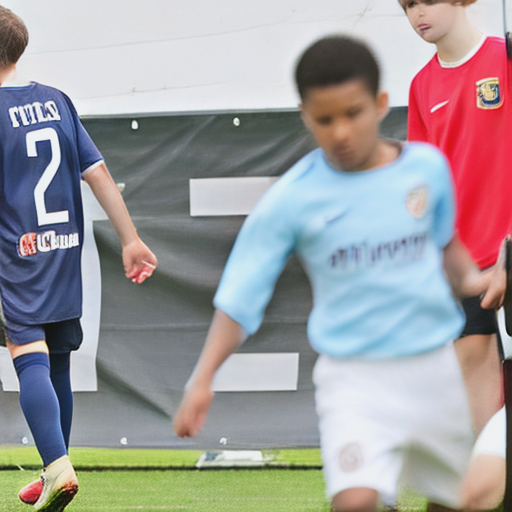} \hspace{-4.5mm} &
    \includegraphics[width=0.09\textwidth]{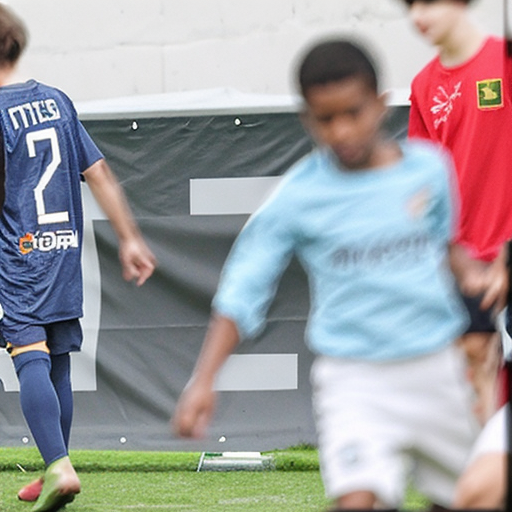} \hspace{-4.5mm} &
    \includegraphics[width=0.09\textwidth]{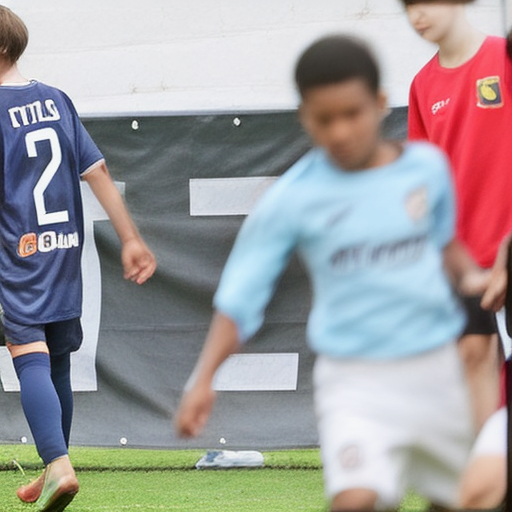} \hspace{-4.5mm} &
    \includegraphics[width=0.09\textwidth]{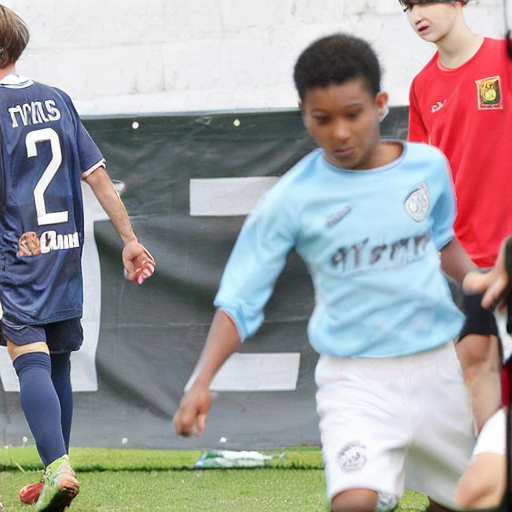} \hspace{-4.5mm} &
    \includegraphics[width=0.09\textwidth]{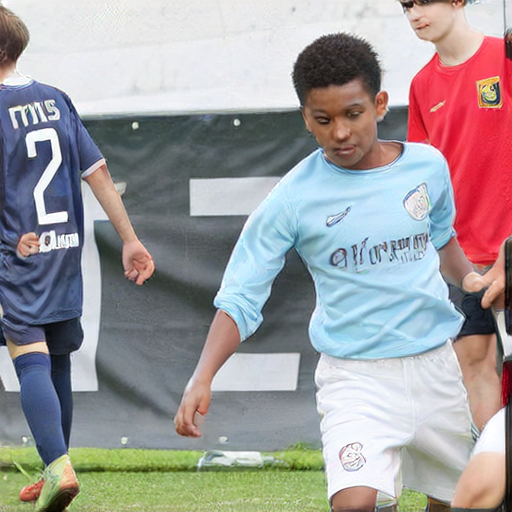} \hspace{-4.5mm} 
    \\
    
    HQ\hspace{-4.5mm} &LQ\hspace{-4.5mm} & SUPIR\hspace{-4.5mm} & SeeSR\hspace{-4.5mm} &PASD\hspace{-4.5mm} &ResShift\hspace{-4.5mm} &OSEDiff\hspace{-4.5mm} &InvSR\hspace{-4.5mm} &OSDHuman\hspace{-4.5mm} &HAODiff\hspace{-4.5mm} &LCUDiff\hspace{-4.5mm} \\
    \end{tabular}
    \end{adjustbox}
}
\end{center}
\vspace{-3mm}
\caption{Visual comparison of the synthetic PERSONA-Val. Please zoom in for a better view.}
\label{fig:vis-val}
\vspace{-3mm}
\end{figure*}

\begin{figure*}[t]

\scriptsize
\begin{center}

\scalebox{0.98}{
    \hspace{-0.4cm}
    \begin{adjustbox}{valign=t}
    \begin{tabular}{cccccccccc}
    \includegraphics[width=0.1\textwidth]{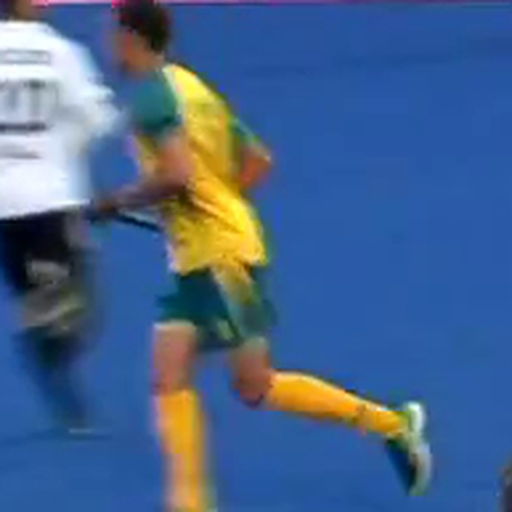} \hspace{-4.5mm} &
    \includegraphics[width=0.1\textwidth]{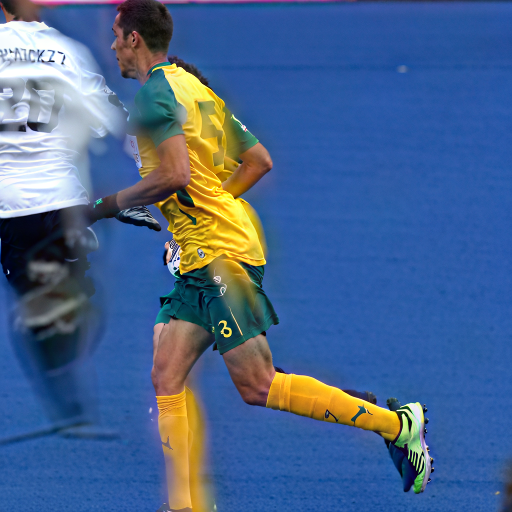} \hspace{-4.5mm} &
    \includegraphics[width=0.1\textwidth]{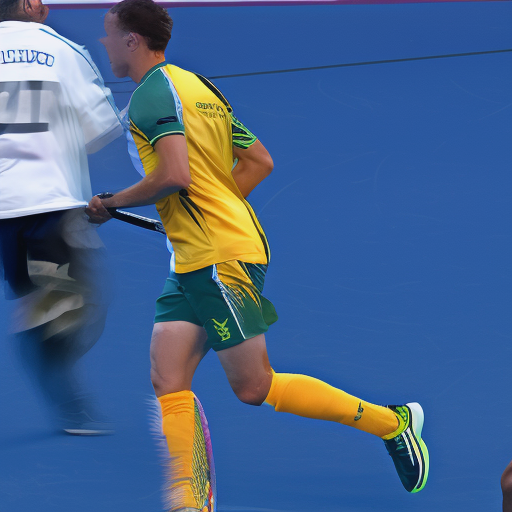} \hspace{-4.5mm} &
    \includegraphics[width=0.1\textwidth]{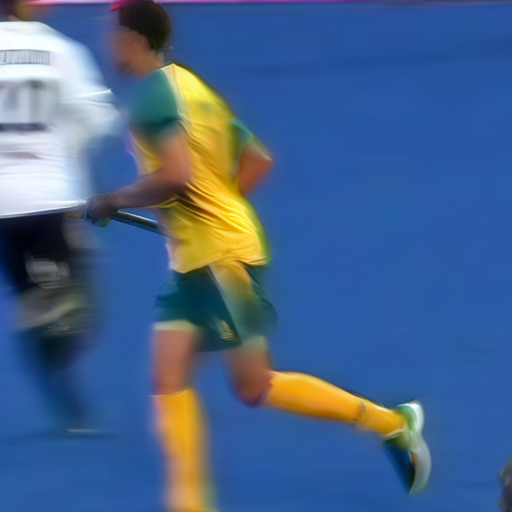} \hspace{-4.5mm} &
    \includegraphics[width=0.1\textwidth]{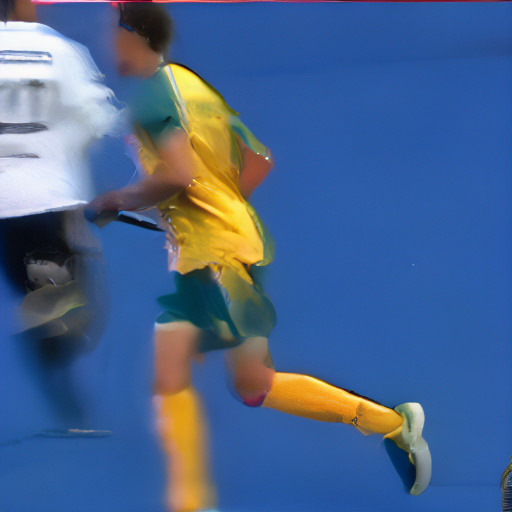} \hspace{-4.5mm} 
    &\includegraphics[width=0.1\textwidth]{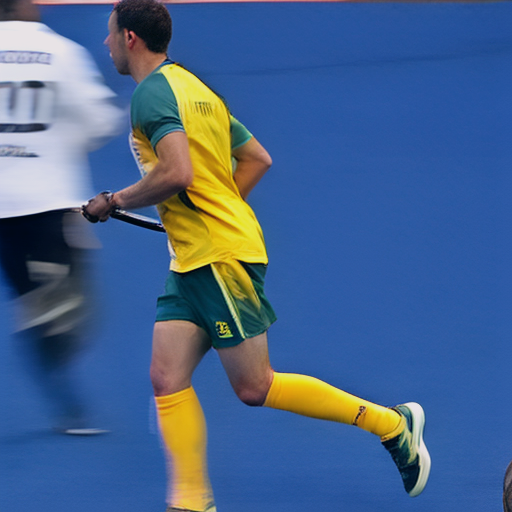} \hspace{-4.5mm} &
    \includegraphics[width=0.1\textwidth]{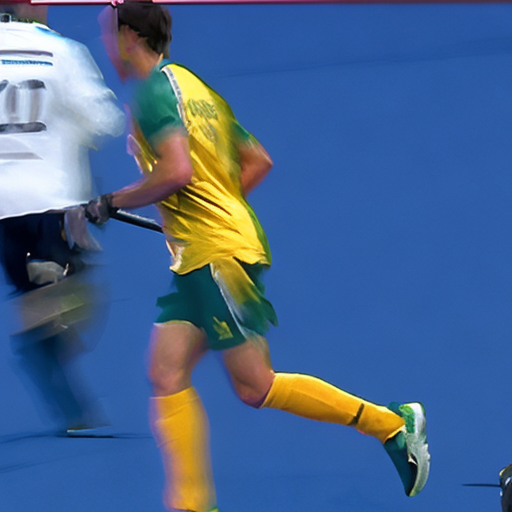} \hspace{-4.5mm} &
    \includegraphics[width=0.1\textwidth]{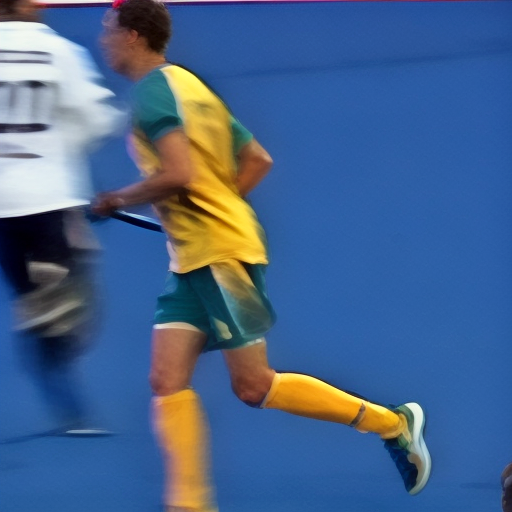} \hspace{-4.5mm} &
    \includegraphics[width=0.1\textwidth]{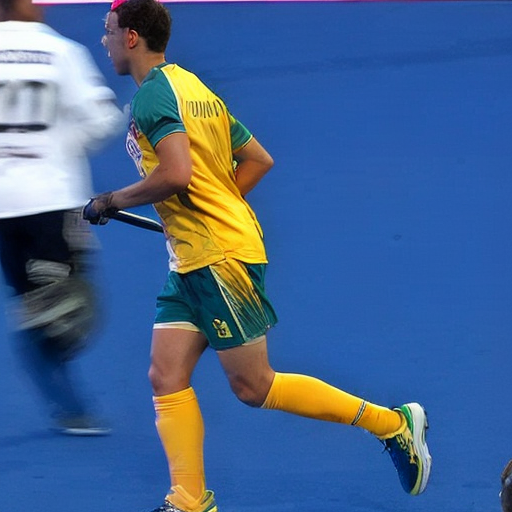} \hspace{-4.5mm} &
    \includegraphics[width=0.1\textwidth]{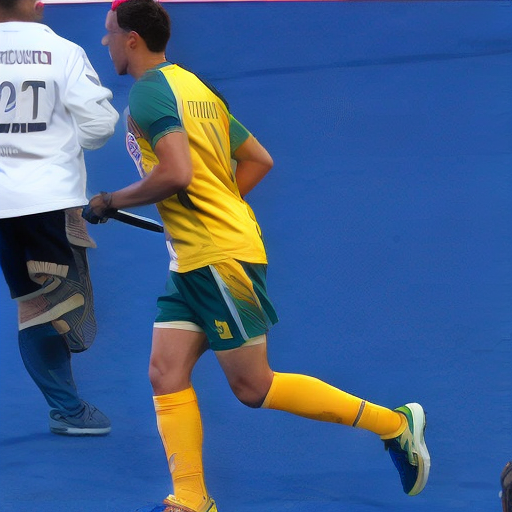} \hspace{-4.5mm} 
    \\
    \end{tabular}
    \end{adjustbox}
    
}
\scalebox{0.98}{
    \hspace{-0.4cm}
    \begin{adjustbox}{valign=t}
    \begin{tabular}{cccccccccc}
    \includegraphics[width=0.1\textwidth]{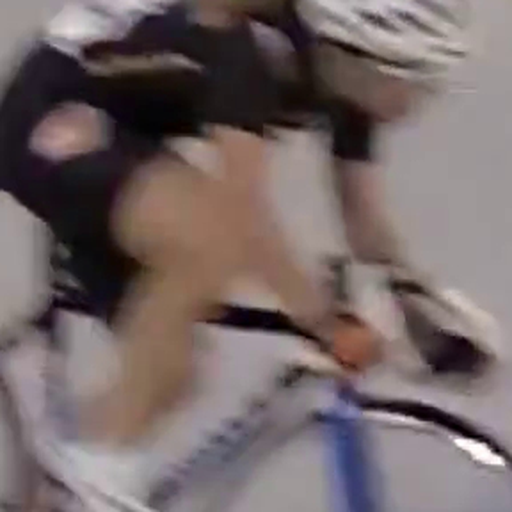} \hspace{-4.5mm} &
    \includegraphics[width=0.1\textwidth]{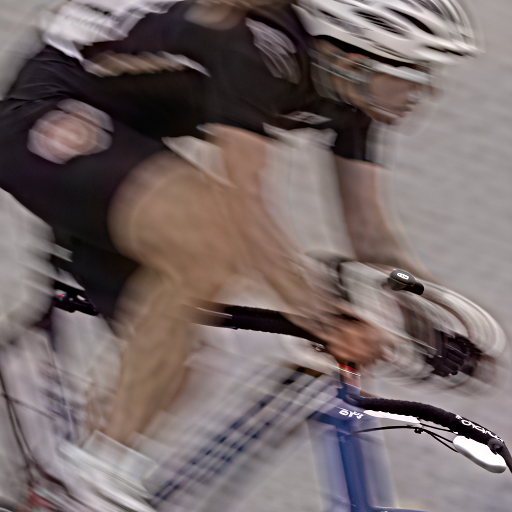} \hspace{-4.5mm} &
    \includegraphics[width=0.1\textwidth]{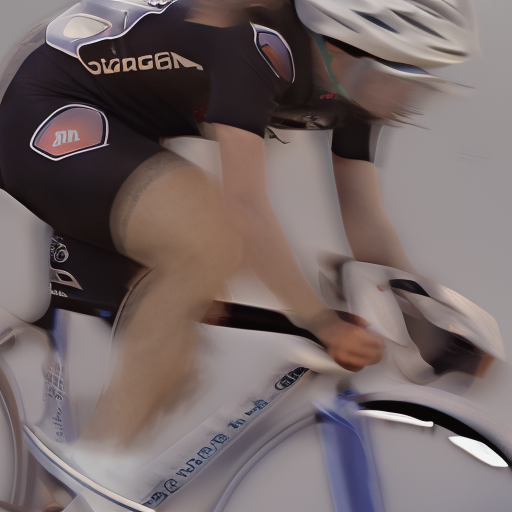} \hspace{-4.5mm} &
    \includegraphics[width=0.1\textwidth]{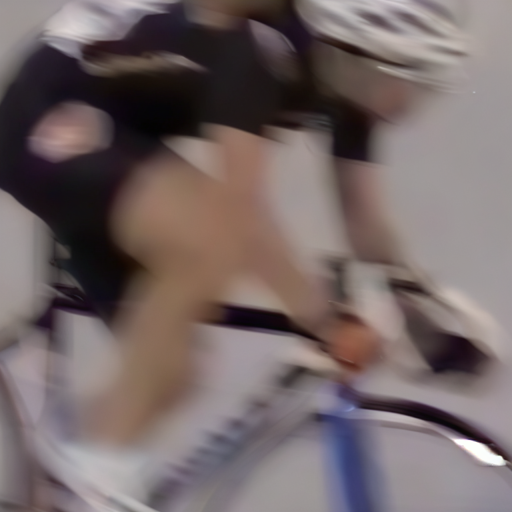} \hspace{-4.5mm} &
    \includegraphics[width=0.1\textwidth]{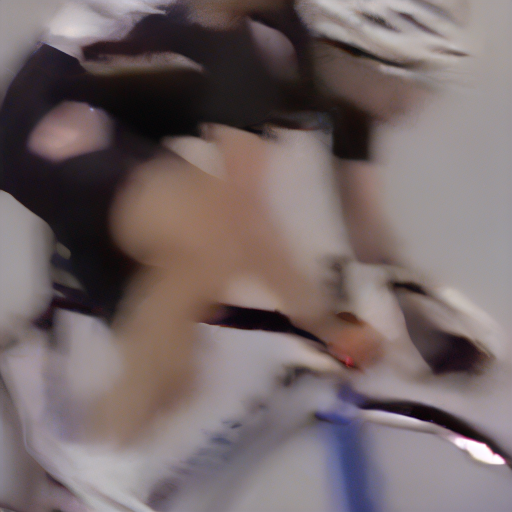} \hspace{-4.5mm} 
    &\includegraphics[width=0.1\textwidth]{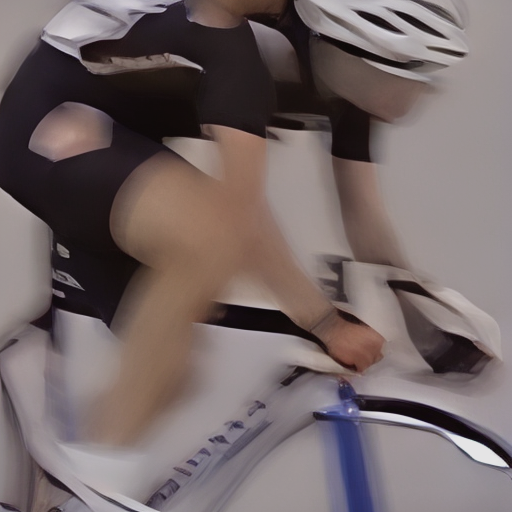} \hspace{-4.5mm} &
    \includegraphics[width=0.1\textwidth]{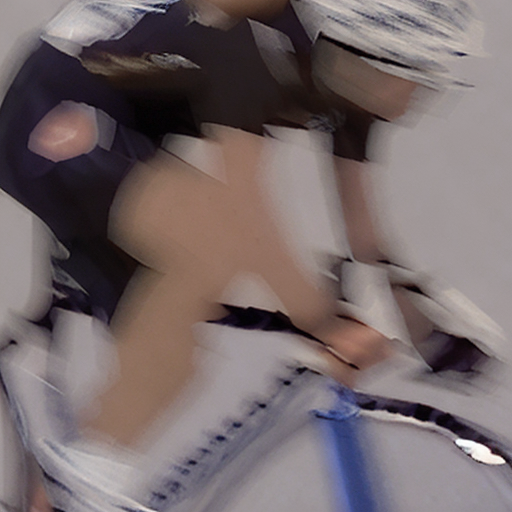} \hspace{-4.5mm} &
    \includegraphics[width=0.1\textwidth]{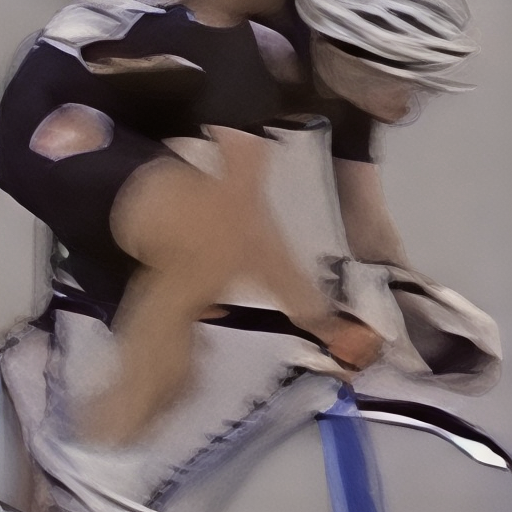} \hspace{-4.5mm} &
    \includegraphics[width=0.1\textwidth]{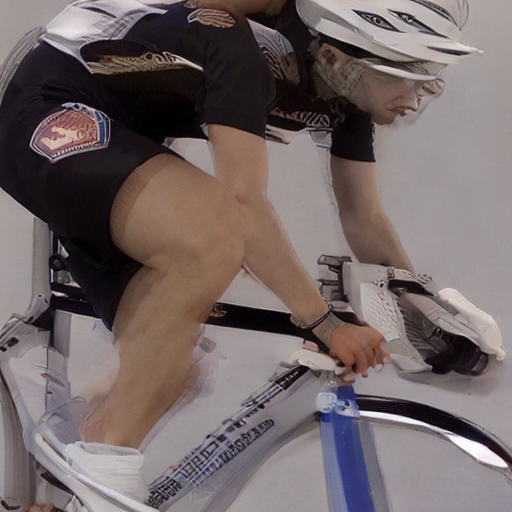} \hspace{-4.5mm} &
    \includegraphics[width=0.1\textwidth]{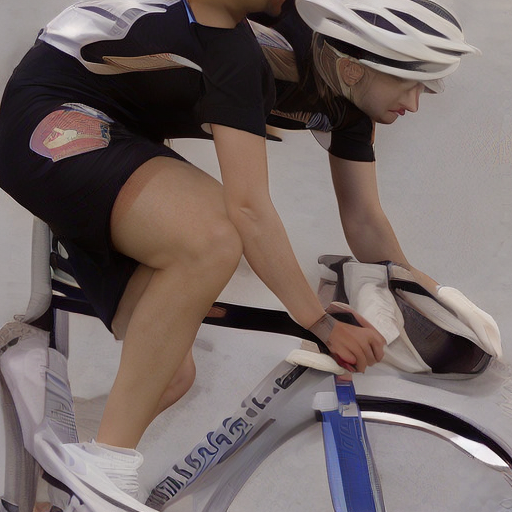} \hspace{-4.5mm} 
    \\
    \end{tabular}
    \end{adjustbox}
    
}
\scalebox{0.98}{

    \hspace{-0.4cm}
    \begin{adjustbox}{valign=t}
    \begin{tabular}{cccccccccc}
    \includegraphics[width=0.1\textwidth]{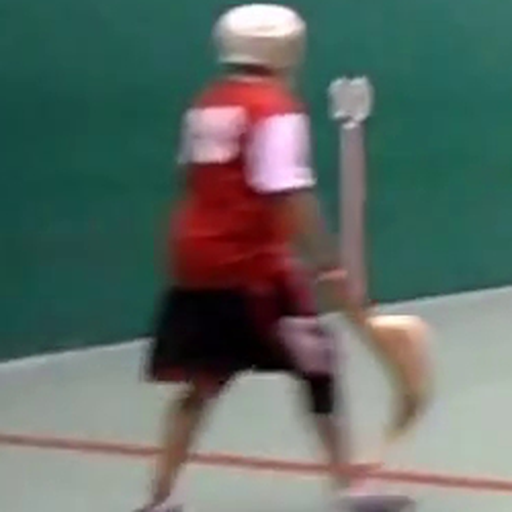} \hspace{-4.5mm} &
    \includegraphics[width=0.1\textwidth]{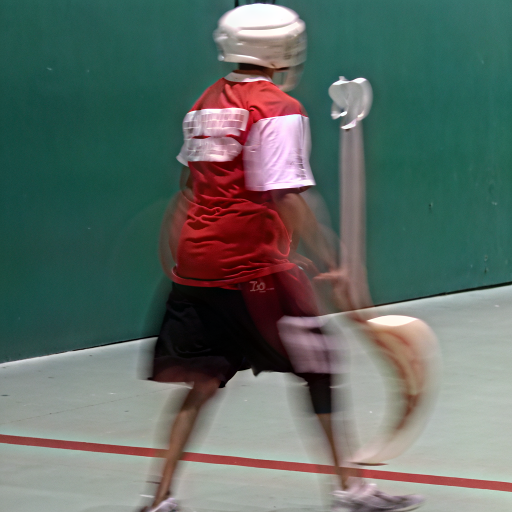} \hspace{-4.5mm} &
    \includegraphics[width=0.1\textwidth]{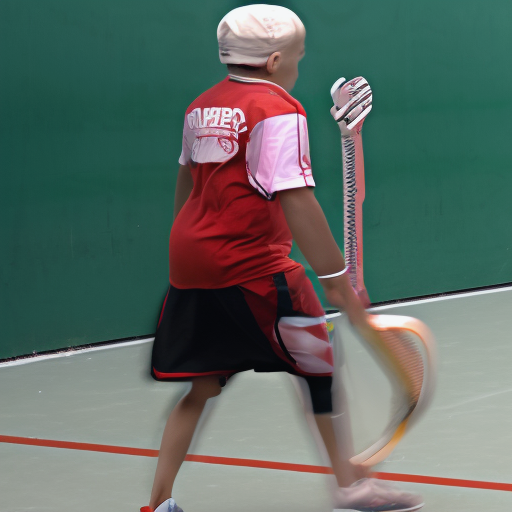} \hspace{-4.5mm} &
    \includegraphics[width=0.1\textwidth]{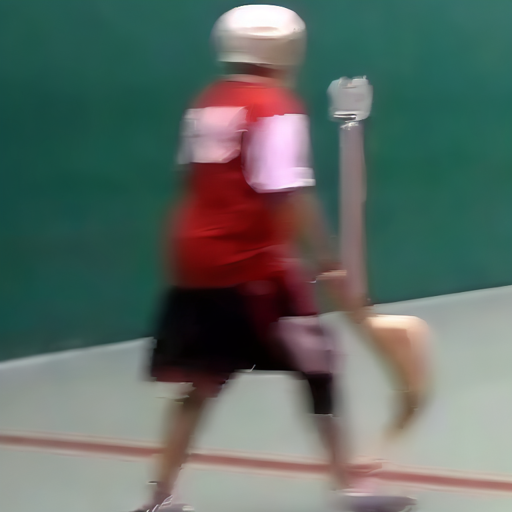} \hspace{-4.5mm} &
    \includegraphics[width=0.1\textwidth]{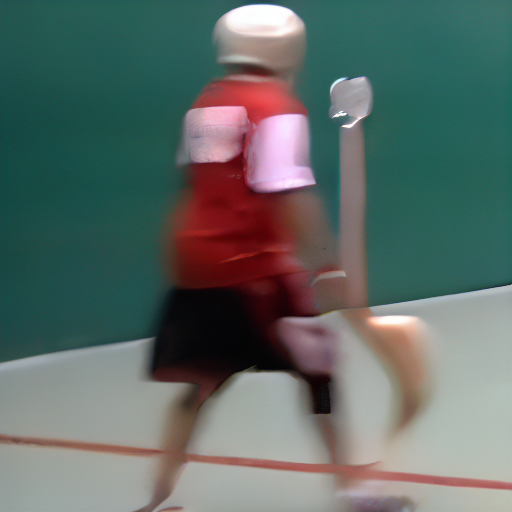} \hspace{-4.5mm} 
    &\includegraphics[width=0.1\textwidth]{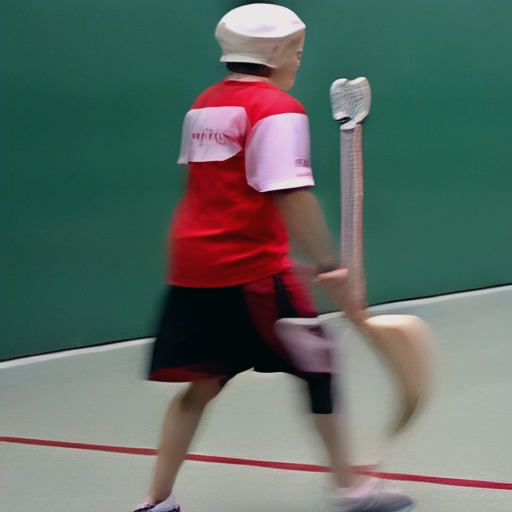} \hspace{-4.5mm} &
    \includegraphics[width=0.1\textwidth]{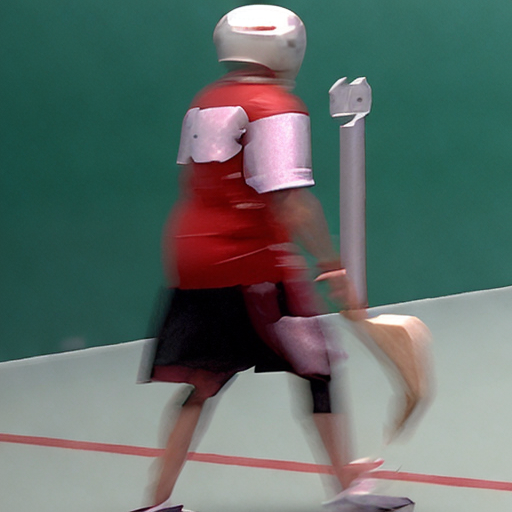} \hspace{-4.5mm} &
    \includegraphics[width=0.1\textwidth]{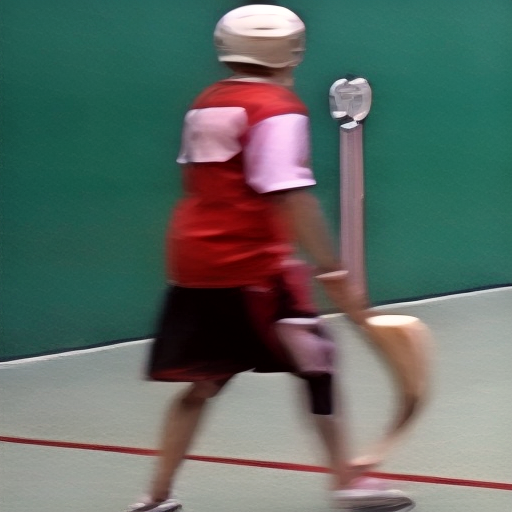} \hspace{-4.5mm} &
    \includegraphics[width=0.1\textwidth]{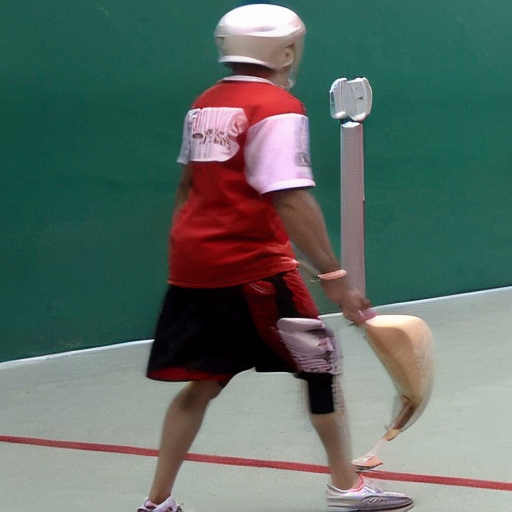} \hspace{-4.5mm} &
    \includegraphics[width=0.1\textwidth]{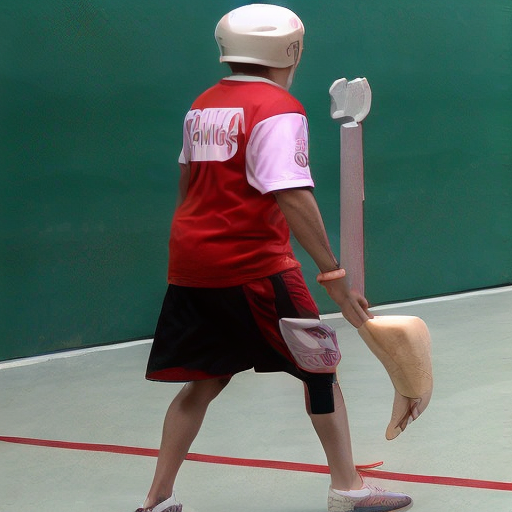} \hspace{-4.5mm} 
    \\
    LQ \hspace{-4.5mm} &
    SUPIR \hspace{-4.5mm} &
    SeeSR \hspace{-4.5mm} &
    PASD \hspace{-4.5mm} &
    ResShift \hspace{-4.5mm} &
    OSEDiff \hspace{-4.5mm} &
    InvSR \hspace{-4.5mm} &
    OSDHuman \hspace{-4.5mm} &
    HAODiff \hspace{-4.5mm} &
    LCUDiff \hspace{-4.5mm} \\
    \end{tabular}
    \end{adjustbox}
    
}
\end{center}

\vspace{-3mm}
\caption{Visual comparison of the real-world MPII-Test dataset in challenging and representative cases. Please zoom in for a better view.}
\label{fig:vis-test}
\vspace{-5mm}
\end{figure*}

\begin{table}[t] 
\scriptsize
\setlength{\tabcolsep}{0.5mm}
\renewcommand{\arraystretch}{1.1}
\centering
\newcolumntype{C}{>{\centering\arraybackslash}X}
\begin{tabularx}{\columnwidth}{l|CCCC}
\toprule
\rowcolor{color3} \textbf{Methods} 
& C\mbox{-}IQA$\uparrow$ & H\mbox{-}IQA$\uparrow$ & TOPIQ$\uparrow$ & TRES$\uparrow$ \\
\midrule
SUPIR~\cite{yu2024supir} & 0.6702 & 0.6405 & 0.6418 & 82.1738 \\ 
DiffBIR~\cite{lin2024diffbir} & 0.6531 & 0.6631 & 0.6449 & 82.7890 \\
SeeSR~\cite{wu2024seesr} & 0.6478 & \textbf{0.6878} & \textbf{0.6883} & \textbf{90.5923} \\ 
PASD~\cite{yang2023pasd} & 0.4023 & 0.4509 & 0.4272 & 54.8259 \\ 
ResShift~\cite{yue2023resshift} & 0.4356 & 0.4824 & 0.4716 & 64.5461 \\ 
\midrule
SinSR~\cite{wang2024sinsr} & 0.4816 & 0.4895 & 0.4710 & 66.4505 \\ 
OSEDiff~\cite{wu2024osediff} & 0.6385 & \textcolor{blue}{0.6214} & \textcolor{blue}{0.5961} & \textcolor{blue}{81.8084} \\
InvSR~\cite{yue2024invsr} & 0.6166 & 0.5798 & 0.5469 & 75.8370 \\ 
OSDHuman~\cite{gong2025osdhuman} & 0.6537 & 0.5975 & 0.5756 & 78.4518 \\ 
HAODiff~\cite{gong2025haodiff} & \textcolor{blue}{0.6923} & 0.5886 & 0.5730 & 78.4821 \\
\midrule   
LCUDiff (ours) & \textcolor{red}{\textbf{0.7076}} & \textcolor{red}{0.6797} & \textcolor{red}{0.6833} & \textcolor{red}{89.8133} \\ 
\bottomrule
\end{tabularx}
\vspace{0.5mm}
\caption{Quantitative results on MPII-Test. Among one-step diffusion methods, the best and second-best are marked in \textcolor{red}{red} and \textcolor{blue}{blue}. The best overall results are in \textbf{bold}. C-IQA and H-IQA denote CLIPIQA and HyperIQA, respectively.}
\vspace{-9mm}
\label{table:mpii_metrics}
\end{table}

\textbf{Implementation Details.}
Our training pipeline consists of three stages. In the first stage, we fine-tune the VAE using Eq.~\eqref{eq:vae_rec} with $\lambda_{p}\!=\!\lambda_{e}\!=\!2$, optimized by AdamW~\cite{loshchilov2018AdamW} with a learning rate of $1\!\times\!10^{-5}$ and a batch size of 8 for 70k iterations on 2 NVIDIA RTX A6000 GPUs (denoted as A6000). In the second stage, we set $\lambda_\text{G}\!=\!0.5$ in Eq.~\eqref{eq:model_loss} and use SDXL~\cite{podell2023sdxl} UNet as the discriminator following D$^3$SR~\cite{li2025d3sr}; the CFG scale $\lambda_\text{cfg}$ is 3.5. We adopt SD-Turbo~\cite{stabilityai_sdturbo_2023} as the base model and train the UNet with LoRA~\cite{hu2022lora} of rank 64 using AdamW with $1\!\times\!10^{-5}$ learning rate and 8 batch size for 60k iterations on 8 A6000 GPUs. In the third stage, we optimize DeR using AdamW with a learning rate of $1\!\times\!10^{-3}$ and a batch size of 64 for 2k iterations on 2 A6000 GPUs.

\textbf{Compared State-of-the-Art (SOTA) Methods.} 
We compare our model against a comprehensive set of leading restoration approaches. The multi-step diffusion methods include SUPIR~\cite{yu2024supir}, DiffBIR~\cite{lin2024diffbir}, SeeSR~\cite{wu2024seesr}, PASD~\cite{yang2023pasd}, and ResShift~\cite{yue2023resshift}, as well as one-step diffusion methods, include SinSR~\cite{wang2024sinsr}, OSEDiff~\cite{wu2024osediff}, InvSR~\cite{yue2024invsr}, OSDHuman~\cite{gong2025osdhuman}, and HAODiff~\cite{gong2025haodiff}.

\begin{table}[t] 
\small
\setlength{\tabcolsep}{0.5mm}
\renewcommand{\arraystretch}{1.1}
\centering
\newcolumntype{C}{>{\centering\arraybackslash}X}
\begin{tabularx}{\columnwidth}{l|CCCC}
\toprule
\rowcolor{color3} \textbf{Methods} 
& DISTS$\downarrow$ & LPIPS$\downarrow$ & PSNRY$\uparrow$ & SSIMY$\uparrow$ \\
\midrule
SinSR & 0.1091 & 0.1784 & 24.71 & 0.7323 \\
OSEDiff & 0.1169 & 0.2031 & 23.00 & 0.6377 \\
InvSR & 0.1205 & 0.2217 & 21.89 & 0.6573 \\
OSDHuman & 0.1462 & 0.2500 & 23.26 & 0.6821 \\
HAODiff & 0.0621 & 0.0966 & 24.98 & 0.7582 \\
\midrule   
LCUDiff (ours) & \textcolor{red}{{0.0620}} & \textcolor{red}{{0.0959}} & \textcolor{red}{{27.40}} & \textcolor{red}{{0.8261}} \\ 
\bottomrule
\end{tabularx}
\vspace{0.5mm}
\caption{Quantitative comparison on light-degradation validation set. The best scores are highlighted in \textcolor{red}{red}. PSNRY and SSIMY denote PSNR and SSIM calculated on the Y channel.}
\vspace{-7mm}
\label{table:additional_metrics}
\end{table}

\subsection{Main Results}

\textbf{Quantitative Comparisons.} Table~\ref{table:model_metrics} presents evaluations on the synthetic PERSONA-Val dataset. Despite the competitive landscape, LCUDiff achieves the best score in DISTS across all methods, highlighting exceptional capability in restoring structure. Moreover, it achieves the highest PSNR and PSNRY among one-step diffusion models, demonstrating that our method strikes a superior balance between inference efficiency and pixel-level fidelity compared to other fast restoration approaches. In evaluations without reference, LCUDiff also secures top ranks in HyperIQA, TOPIQ, and TReS among one-step models, suggesting improved perceptual quality that better correlates with human judgments. Extending this evaluation to real-world scenarios, Tab.~\ref{table:mpii_metrics} reports performance on MPII-Test. LCUDiff consistently outperforms all one-step competitors, leading in metrics such as CLIPIQA and TReS. This suggests that our model generalizes well to real-world human-centered images. It also shows strong robustness various noise and complex motion blur in real-world environments. Crucially, we validate the restoration fidelity under a setting of light degradation in Tab.~\ref{table:additional_metrics}. Our model exhibits a clear and consistent advantage, outperforming the recent one-step baseline by a substantial margin in both pixel-level and perceptual metrics. Since light degradations place a higher emphasis on faithful preservation, this improvement suggests that LCUDiff reduces over-correction and mitigates hallucination artifacts often observed in diffusion models. It better preserves the original subject identity and structural integrity.

\textbf{Qualitative Comparisons.}
Comprehensive visual comparisons are shown in Fig.~\ref{fig:vis-val} and Fig.~\ref{fig:vis-test} on synthetic and real-world datasets, respectively. Across diverse scenarios, LCUDiff consistently delivers clearer and more faithful faces, outperforming competitors under both mild and severe degradations. It also better preserves semantic fidelity and original materials. For example, in Fig.~\ref{fig:vis-val}, multi-step methods such as SUPIR~\cite{yu2024supir}, SeeSR~\cite{wu2024seesr}, and ResShift~\cite{yue2023resshift} fail to recover the blue knitted sweater and instead shift it to smooth or denim-like textures, while OSDHuman~\cite{gong2025osdhuman} avoids strong texture shifts but loses facial details. In contrast, LCUDiff reconstructs the knitted pattern while producing sharper and more natural facial features. Moreover, LCUDiff respects optical realism by preserving natural bokeh and depth consistency, avoiding the over-sharpened backgrounds and cutout artifacts often observed in InvSR~\cite{yue2024invsr}. On MPII-Test (Fig.~\ref{fig:vis-test}), LCUDiff generalizes well to real-world motion blur and achieves more coherent structures in challenging dynamic scenes, while avoiding the excessive smoothing of OSDHuman and reducing residual blur artifacts seen in HAODiff~\cite{gong2025haodiff}. Additional visual comparisons under lighter degradations are provided in the supplementary material. They show that LCUDiff preserves structures in less blurred regions.

\begin{figure}[t]

\scriptsize
\begin{center}

    \begin{adjustbox}{valign=t}
    \begin{tabular}{ccc}
    \hspace{-0.4cm} \includegraphics[width=0.15\textwidth]{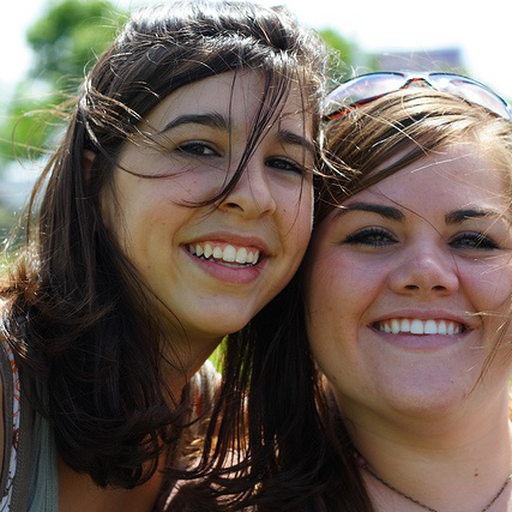} \hspace{-3.5mm} &
    \includegraphics[width=0.15\textwidth]{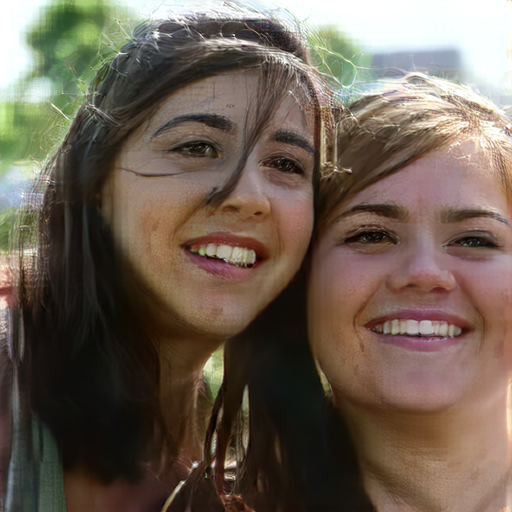} \hspace{-3.5mm} &
    \includegraphics[width=0.15\textwidth]{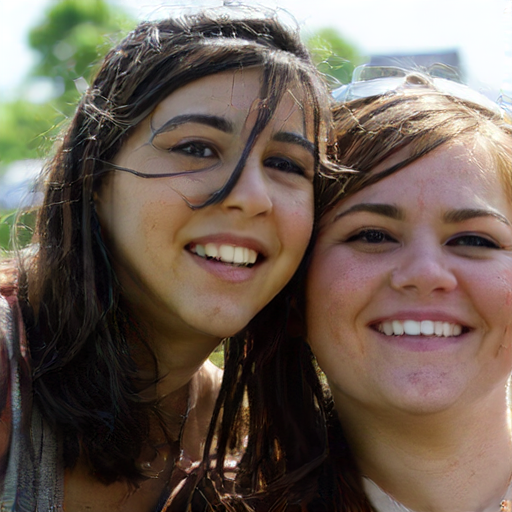} \hspace{-3.5mm} 
    \\
    
    HQ\hspace{-3.5mm} &w/o CSD\hspace{-3.5mm} &w/ CSD\hspace{-3.5mm} \\
    \end{tabular}
    \end{adjustbox}
\end{center}
\vspace{-3.5mm}
\caption{Effect of channel splitting distillation (CSD). Without CSD, the restored result exhibits regular grid-like artifacts due to latent misalignment, whereas CSD stabilizes training and yields a cleaner restoration closer to the HQ target.}
\label{fig:wether_CSD}
\vspace{-6mm}
\end{figure}

\vspace{-2mm}
\subsection{Ablation Study}
\vspace{-1mm}
\textbf{Effect of Channel Splitting Distillation (CSD).}
We conduct an ablation study to evaluate the impact of the CSD loss. For the 16-channel VAE, we train two variants under identical settings, differing only in whether CSD is applied. Each VAE is used to train the subsequent diffusion model. We observe that removing CSD slows down convergence during VAE adaptation. It also increases the risk of gradient explosion during training, which may cause the restoration model to fail. We select a checkpoint trained without CSD that does not diverge and compare it with the CSD-based model in Fig.~\ref{fig:wether_CSD}. As shown, the model trained without CSD produces regular grid-like artifacts. We attribute this issue to a large mismatch between the upgraded latent space and the pretrained prior, which is difficult to bridge through fine-tuning without the alignment enforced by CSD.

\textbf{Effect of VAE Channel Capacity.}
As shown in Tab.~\ref{table:VAE_channel}, we fine-tune VAEs with latent channels, which correspond to different information capacities of the latent space. Increasing the channel count raises the upper bound of preserved latent information. As a result, the full-reference metrics DISTS and LPIPS improve with larger channel capacity. Moreover, with a more informative latent representation, the UNet can better leverage the available details to produce visually higher-quality outputs. This is also reflected by the no-reference metrics, where CLIPIQA and HyperIQA increase as the number of latent channels grows.

\begin{table}[t]
\centering
\small
\setlength{\tabcolsep}{0.5mm}
\renewcommand{\arraystretch}{1.1}
\centering
\newcolumntype{C}{>{\centering\arraybackslash}X}
\begin{tabularx}{\columnwidth}{l|CCCC}
\toprule
\rowcolor{color3} \textbf{Channels} & \textbf{DISTS}$\downarrow$ & \textbf{LPIPS}$\downarrow$ & \textbf{CLIPIQA}$\uparrow$ & \textbf{HyperIQA}$\uparrow$ \\
\midrule
4ch  & 0.1057 & 0.2274 & 0.7176 & 0.6683 \\
8ch  & 0.1035 & 0.2203 & 0.7246 & 0.6675 \\
16ch & \textbf{0.1019} & \textbf{0.2155} & \textbf{0.7432} & \textbf{0.6728} \\
\bottomrule
\end{tabularx}
\caption{Effect of VAE latent channel capacity on PERSONA-Val. Increasing the number of latent channels improves both full-reference and no-reference metrics. Best results are in \textbf{bold}.}
\vspace{-5.5mm}
\label{table:VAE_channel}
\end{table}

\begin{table}[t]
\centering
\small
\setlength{\tabcolsep}{0.5mm}
\renewcommand{\arraystretch}{1.1}
\centering
\newcolumntype{C}{>{\centering\arraybackslash}X}
\begin{tabularx}{\columnwidth}{l|CCCC}
\toprule
\rowcolor{color3}\textbf{Decoding} & \textbf{DISTS}$\downarrow$ & \textbf{LPIPS}$\downarrow$ & \textbf{CLIPIQA}$\uparrow$ & \textbf{HyperIQA}$\uparrow$ \\
\midrule
$\mathcal{D}_{16ch}$ & 0.1008 & 0.2138 & 0.7470 & 0.6643 \\
$\mathcal{D}_{4ch}$  & 0.1141 & 0.2238 & 0.7990 & 0.7298 \\
w/ DeR & 0.1022 & 0.2149 & 0.7540 & 0.6735 \\
\bottomrule
\end{tabularx}
\caption{Decoding strategy comparison on PERSONA-Val. 
$\mathcal{D}_{16ch}$ decodes the full restored latent $\hat{z}_H$ with the fine-tuned 16-channel decoder. 
$\mathcal{D}_{4ch}$ decodes the anchor channels $\hat{z}_{anc}$, i.e., the first four channels of $\hat{z}_H$, using the pretrained SD decoder. 
DeR performs hard per-sample routing between $\mathcal{D}_{16ch}$ and $\mathcal{D}_{4ch}$.}
\label{table:decoder_der}
\vspace{-6.5mm}
\end{table}

\textbf{Decoder Choice and Decoder Router (DeR).}
DeR routes between the pretrained decoder and the fine-tuned decoder, but it is optional. As shown in Tab.~\ref{table:decoder_der}, using either decoder alone already yields reasonable performance. The two decoders exhibit complementary tendencies. $\mathcal{D}_{4ch}$ tends to produce higher no-reference scores, while $\mathcal{D}_{16ch}$ achieves better full-reference fidelity. This complementarity suggests that the preferred decoder can vary across images and degradation regimes. However, relying solely on a single decoder can lead to an imbalanced outcome on individual images. DeR mitigates this issue by choosing the decoding path that better matches the current input, reducing metric trade-offs in practice. DeR therefore performs per-sample routing to improve the trade-off between fidelity and perceptual quality across diverse conditions. Further analysis is provided in the supplementary material.

\vspace{-2mm}
\section{Conclusion}
\vspace{-1mm}
We presented LCUDiff, an efficient one-step diffusion framework for faithful human body restoration. Our key idea is to upgrade the latent capacity of a pretrained latent diffusion model while preserving its generative prior. We expand the VAE latent space from 4 to 16 channels and introduce channel splitting distillation to align the anchor channels with the pretrained latent space, while allocating extra channels specifically for high-frequency details. To bridge the channel mismatch with the pretrained diffusion U-Net, we propose prior-preserving adaptation with a dual-branch input design and a smooth fusion schedule, which stabilizes training without increasing inference overhead. We also introduce a decoder router for per-sample decoding selection, leveraging the complementary behavior of the original and fine-tuned decoders. Extensive experiments on both synthetic and real-world benchmarks demonstrate that LCUDiff improves both pixel-level and perceptual fidelity, reduces artifacts across diverse degradation levels, and maintains the efficiency of one-step inference.

\bibliography{main}
\bibliographystyle{icml2026}

\end{document}